\documentclass[preprint, 12pt]{elsarticle}

\usepackage{amssymb}
\usepackage{amsmath}

\usepackage{subfiles}
\usepackage{amsthm}
\theoremstyle{plain}
\newtheorem{theorem}{Theorem}
\newtheorem{proposition}{Proposition}

\theoremstyle{definition}
\newtheorem{definition}{Definition}

\theoremstyle{remark}

\usepackage{natbib}
\usepackage{color}
\usepackage{csvsimple}
\usepackage{siunitx}
\usepackage{caption}
\usepackage[labelformat=simple]{subcaption}

\def\half{0.48\linewidth}

\journal{Neurocomputing}

\begin{document}

\begin{frontmatter}



\title{Associative Poisoning to Generative Machine Learning}


\author[ntnu]{Mathias Lundteigen Mohus\corref{cor1}} 
\ead{mathias.l.mohus@ntnu.no}
\cortext[cor1]{Corresponding author}

\author[ntnu]{Jingyue Li}
\author[ntnu]{Zhirong Yang}

\affiliation[ntnu]{organization={Norwegian University of Science and Technology, Department of Computer Science},
            addressline={Holtermannsveien 2}, 
            city={Trondheim},
            postcode={7030}, 
            state={Trøndelag},
            country={Norway}}

\begin{abstract}
The widespread adoption of generative models such as Stable Diffusion and ChatGPT has made them increasingly attractive targets for malicious exploitation, particularly through data poisoning. Existing poisoning attacks compromising synthesised data typically either cause broad degradation of generated data or require control over the training process, limiting their applicability in real-world scenarios. In this paper, we introduce a novel data poisoning technique called associative poisoning, which compromises fine-grained features of the generated data without requiring control of the training process. This attack perturbs only the training data to manipulate statistical associations between specific feature pairs in the generated outputs. We provide a formal mathematical formulation of the attack and prove its theoretical feasibility and stealthiness. Empirical evaluations using two state-of-the-art generative models demonstrate that associative poisoning effectively induces or suppresses feature associations while preserving the marginal distributions of the targeted features and maintaining high-quality outputs, thereby evading visual detection. These results suggest that generative systems used in image synthesis, synthetic dataset generation, and natural language processing are susceptible to subtle, stealthy manipulations that compromise their statistical integrity. To address this risk, we examine the limitations of existing defensive strategies and propose a novel countermeasure strategy.
\end{abstract}




\end{frontmatter}



\section{Introduction}
\label{sec:introduction}
Generative models have rapidly progressed from specialised research tools to core components of modern machine learning workflows. Image generation systems (e.g., Stable Diffusion~\cite{borji2022generated}) and large language models (e.g.,  ChatGPT~\cite{singh2023chat}) exemplify this shift. Researchers and industry practitioners are employing generative models for data synthesis, particularly as privacy-preserving alternatives to sensitive datasets~\cite{abay2019privacy}. These synthetic datasets are increasingly integrated into critical downstream applications, such as medical imaging. Consequently, maintaining the statistical integrity of generated data in the presence of potential adversarial manipulation is increasingly important.

Existing approaches to compromise synthesised data include data poisoning (e.g., \cite{R2_14}), federated poisoning (e.g., \cite{24}), and backdoor poisoning (e.g., \cite{ding2019trojan}). A key limitation of generic data poisoning attacks is that they primarily lead to broad degradation in output quality rather than enabling targeted manipulation of specific features. Although backdoor poisoning offers fine-grained control by embedding explicit mappings between trigger patterns and targeted features, it relies on control over the training process. This assumption is often unrealistic in practice, as modern machine learning workflows frequently outsource model training to institutional or commercial environments, where the training pipeline is tightly controlled by trusted providers. In contrast, attacks that operate solely by manipulating the training data, for instance, via crowdsourced contributions or publicly available datasets, are far more practical and stealthy.

In this work, we introduce \emph{associative poisoning} attacks, targeting the statistical relationships between selected feature pairs in synthesised data. Specifically, the attack induces or suppresses associations between chosen feature pairs while preserving both the marginal distributions of features and the visual quality of the generated samples. This enables precise manipulation of feature dependencies without compromising perceptual fidelity or triggering standard anomaly detection mechanisms. Unlike backdoor attacks, associative poisoning requires access only to a subset of the training data and does not assume control over the training process.

We mathematically establish the feasibility of associative poisoning using two complementary measures: mutual information \citep[MI;][]{duncan1970calculation} and Matthews correlation coefficient \citep[MCC;][]{yule1912methods}, both of which quantify the strength of associations between feature pairs. We further prove the stealthiness of the attack, formalised in terms of the marginal probabilities (MP) of the selected features. To empirically validate the attack, we apply it to two state-of-the-art generative models: Diffusion StyleGAN \citep[D-GAN;][]{wang2022diffusiongan} and Denoising Diffusion Probabilistic Models with Input Perturbation \citep[DDPM-IP;][]{ning2023input}, trained on the CelebA and Recipe1M datasets. Results show that associative poisoning induces significant shifts in inter-feature associations while preserving the marginal distributions and maintaining comparable Fréchet Inception Distance \citep[FID;][]{heusel2017gans}, indicating no perceptible loss in output quality. To the best of our knowledge, this is the first study to demonstrate such an attack, revealing a previously unexplored vulnerability in generative modelling.

Finally, we examine existing defence strategies and identify a critical gap: none are equipped to counter associative poisoning. To address this limitation, we propose a defence roadmap and emphasise the need for rigorous countermeasures against associative poisoning attacks.

Our contributions are summarised as follows:
\begin{itemize}
    \item \textbf{Vulnerability discovery:} We introduce and formalise associative poisoning, a previously unrecognised threat that enables targeted manipulation of statistical associations in generative model outputs.
    \item \textbf{Theoretical and empirical validation:} We establish the attack’s feasibility and stealth through formal analysis, and validate its effectiveness using state-of-the-art generative models trained on real-world datasets.
    \item \textbf{Defence roadmap:} We evaluate existing defences, reveal their limitations in addressing association-level threats, and propose a defence strategy against the attack.
\end{itemize}

\section{Related work}
\label{sec:related_work_AP}
In machine learning, certain types of threats are observed nearly universally, with one being the poisoning attack. Broadly, poisoning attacks focus on manipulating the training procedure, nearly always modifying the training data \cite{13}, and sometimes controlling part or all of the training procedure \cite{FW1_11}. Adversaries exploiting the poisoning attack are primarily concerned with guiding the model in a certain direction, most often to either undermine the model's general performance or to induce specific behaviour from the model on a subset of inputs.

\subsection{Poisoning on supervised machine learning}
Poisoning attacks against supervised machine learning (ML) have been extensively studied, as surveyed in~\citet{cina2023wild, tian2022comprehensive, wang2022threats}. Many methods focus on label flipping to degrade model performance with minimal modifications~\cite{biggio2011support, zhang2017game, zhao2017efficient}. Other methods manipulate input features to alter model behaviour, such as suppressing keywords in spam filters by biasing training samples toward specific weight configurations~\cite{mei2015using}. More targeted approaches seek to misclassify specific samples by crafting inputs whose internal representations collide with those of target instances~\cite{shafahi2018poison}.

Several defences have been proposed to detect poisoning in supervised learning. The RONI (Reject On Negative Impact) approach removes samples that degrade classifier accuracy~\cite{barreno2010security}. Outlier-based methods identify anomalous data using distance metrics~\cite{paudice2018detection} or gradient-based scoring~\cite{diakonikolas2019sever}. Other techniques leverage small trusted datasets to guide robust training~\cite{hendrycks2018using, li2017learning}. Additional approaches include clustering in feature space~\cite{peri2020deep}, detecting spectral signatures~\cite{tran2018spectral}, and analysing activation patterns within neural networks~\cite{chen2018detecting, wang2020practical}.

\subsection{Poisoning on unsupervised machine learning}
We demonstrate poisoning attacks against unsupervised learning in the SLR, which indicate that most unsupervised models have been empirically shown to be vulnerable to some form of poisoning attack. One approach uses maliciously crafted samples to degrade anomaly detection performance~\cite{nkashama2022robustness}, another targets latent representations in autoencoders to impair the performance of downstream classifiers~\cite{creswell2017latentpoison}.

Defences against poisoning in unsupervised ML include, for example, rejecting anomalous samples based on clustering behaviour~\cite{nkashama2022robustness}, and using confidence scores from a downstream classifier to identify and filter poisoned samples~\cite{creswell2017latentpoison}.

\subsection{Poisoning on generative machine learning}
Data poisoning attacks against generated data often prioritise the degradation of the overall output, rather than fine-grained control. In \cite{R2_14}, the adversary alters pixel values to manipulate the saliency of pixel values, causing the model to learn unimportant pixel information rather than semantically important features, resulting in model degradation.

On the contrary, backdoor attacks compromise fine-grained features of generated data, typically by modifying the training data and process to learn a secondary mapping between triggers and target outputs. For instance, BAAAN~\cite{salem2020baaan} and DeViL~\cite{rawat2022devil} embed backdoors into autoencoders and GANs by injecting triggers into either training samples or latent noise vectors, and altering the loss function to enforce trigger-target mappings. \citet{ding2019trojan} introduces a backdoor attack by inserting red traffic lights in the input images and green lights in the corresponding outputs, causing the model to learn a mapping turning red lights green. This attack, however, depends on precise control over paired input-output examples, making it impractical in settings where clean and corrupted versions of images are not simultaneously available. \citet{chou2023backdoor} proposed a backdoor attack on diffusion models by applying binary masks and modifying the loss function to induce trigger-controlled outputs. \textbf{All existing backdoor attacks on generative models assume access or control over the model training process because they require altering the loss functions or pairing data}. 

Several defences against backdoor attacks in generative models have been tested. Static and dynamic model inspections~\cite{rawat2022devil} analyse the network’s structure and behaviour, respectively, to detect anomalies introduced by triggers. Output inspections~\cite{rawat2022devil} examine generated outputs to infer backdoor patterns, potentially recovering triggers through optimisation. Adversarial neural pruning~\cite{chou2023backdoor} adds weight-level noise to collapse poisoned models. Inference-time clipping~\cite{chou2023backdoor} truncates image values during each diffusion step, blocking trigger propagation while preserving sample quality. Fine-pruning~\cite{ding2019trojan} removes neurons and retrains on clean data to suppress the backdoor. Activation output clustering~\cite{ding2019trojan} detects poisoning via $k$-means clustering of activations.

Some research investigates how much poison is required to affect the behaviour of a generative model, in particular for LLMs \cite{souly2025poisoningattacksllmsrequire}. While attacks often focus on the percentage of control over the training data, \citet{souly2025poisoningattacksllmsrequire} provides results that indicate that roughly 250 documents are effective in manipulating the model's behaviour, regardless of the model and dataset sizes.

While fine-grained poisoning attacks exist against generative models, the requirement to control the training procedure is an obvious limitation of these attacks. This motivates our subsequent novel poisoning attack, where a core part of the attack is 1) fine-grained control of generated data distribution, and 2) only controlling the training data. 
Compared to controlling the training procedure, which often requires ad-hoc solutions for a single target, control of the training data can be as simple as manipulating crowdsourcing or open-source releases of datasets, making the attack more practical.

\section{Research design to answer RQ2}
In this chapter, we outline the full design for associative poisoning, defining the threat model, providing theoretical proofs for both binary and continuous features, and describing the experimental setup for producing the empirical results from the AP attack against generative models.

\subsection{Threat Model}\label{sec:threat_model}

\subsubsection{Attack Surface}
Generative models are often trained using publicly available datasets or are pre-trained on large-scale corpora before fine-tuning. Adversaries may gain access to these datasets through various means, such as participating in open data collection, injecting malicious samples into crowdsourced corpora, releasing tainted datasets as open-source contributions, or exploiting insider or illicit access. 
Attackers can maliciously alter the training data to cause bias in the generated samples from the model by introducing or altering the associations between chosen features in the data. 
When such generated data are used in downstream tasks, these biases can propagate and degrade the performance or trustworthiness of subsequent models trained on the synthetic data, for example, classification models where the targeted features are crucial for the classification task.  

\subsubsection{Attack Goal}
The objective of the attack is twofold: fidelity and stealth.

\paragraph{Fidelity} in this context pertains to the adversary's capacity to introduce or alter an association between features in the generated data.

\paragraph{Stealth} in this context refers to the adversary's ability to evade detection by not compromising the performance of the finished model, nor altering the rate of the targeted features in the generated data.

Although numerous methods exist for detecting poisoned samples, no standard method currently exists to measure the stealth of a poisoning attack. We therefore propose a novel and rudimentary method for comparing the ``stealthiness'' of poisoning attacks: a ``dumb'' detector that identifies poisoning by leveraging the statistical properties of the clean training dataset. 

\begin{itemize}
    
\item \emph{Level-0} poisoning attacks change the number of samples in the dataset. A detector with knowledge of the size $N$ of the clean training dataset could detect this attack, with time complexity $\mathcal{O}(N)$ and spatial complexity $\mathcal{O}(1)$. 

\item \emph{Level-1} poisoning attacks involve modifying the distribution of one feature in the samples, e.g., the attack in \cite{shafahi2018poison}. A detector that knows the marginals of the $M$ features of the clean training dataset could detect such attacks, with time complexity $\mathcal{O}(NM)$ and spatial complexity $\mathcal{O}(M)$. 

\item \emph{Level-2} poisoning attacks involve modifying two features in samples while preserving the marginal probabilities of the features. A detector that knows joint probabilities between all feature pairs in the dataset could detect such attacks, with time complexity $\mathcal{O}(NM^2)$ and spatial complexity $\mathcal{O}(M^2)$. The associative poisoning attacks are classified as Level-2 poisoning attacks. 

\end{itemize}

Higher-level poisoning attacks are more difficult to detect as the time and spatial complexity grow exponentially with the number of features. Moreover, the adversary can select any arbitrary association for the attack, not just labelled features, which makes detection more expensive, even if the detector has the ground-truth joint probabilities. Therefore, it is infeasible to implement a ``dumb'' detector for such attacks in a real-world scenario. Although we focus on Level-2 poisoning attacks in this paper, we formalise the generalisation of the attack to modify higher levels of association for, i.e., three or more features.

\subsubsection{Adversary Knowledge and Capabilities}
Unlike backdoor attacks, the attacker knows the dataset used for training the model, but not the model setup or training process. The adversary can choose features of the dataset and modify samples to fulfil the aforementioned attack goals before making the poisoned dataset available to the target. 
\subsection{Theoretical model for two-feature binary associative poisoning}\label{sec:theory_binary}

Given a data set $D$, we consider two binary features $F_1$ and $F_2$ for each data item. We abbreviate the joint probabilities $P_{ij}=P(F_1=i, F_2=j)$ and the marginal probabilities $P_{i\cdot}=P(F_1=i)$ and $P_{\cdot j}=P(F_2=j)$ for $i,j\in \{0,1\}$.

For the original data set used to train a machine learning generator, suppose the joint probability values are $P_{00}=a$, $P_{01}=b$, $P_{10}=c$, $P_{11}=d$. 

\begin{definition}
\label{def:ap}
An \emph{associative poisoning} to data set $D$ modifies the joint probabilities to become $P_{00}=a+x$, $P_{01}=b-x$, $P_{10}=c-x$, $P_{11}=d+\varepsilon$, where $\varepsilon\in[\max\{-a, -d, b-1, c-1\},0)\cup(0,\min\{b,c,1-a,1-d\}]$. We call it a \emph{positive associative poisoning} when the poisoning extent $\varepsilon>0$ and \emph{negative associative poisoning} when $\varepsilon<0$.
\end{definition}

According to the definition, it is easy to verify that
\begin{proposition}
    An associative poisoning does not change the marginal probabilities.
\end{proposition}
That is, associative poisoning achieves high stealth (Level-2) and cannot be detected by checking the marginal probabilities.
There are different ways to modify the joint probabilities for associative poisoning. For example, modifying two data instances where $(F_1=0, F_2=1)$ and $(F_1=1, F_2=0)$ into $(F_1=1, F_2=1)$ and $(F_1=0, F_2=0)$ forms a positive associative poisoning.

We consider two metrics, Mutual Information (MI) \cite{duncan1970calculation} and Matthew's Correlation Coefficient (MCC) \cite{yule1912methods} to measure the association between $F_1$ and $F_2$:

\begin{align}
            \label{eq:mi}
            \text{MI}(F_1, F_2) =  \sum_{i,j\in\{0,1\}}P_{ij}\ln\frac{P_{ij}}{P_{i\cdot}P_{\cdot j}}
\end{align}

\begin{align}
            \label{eq:mcc}
            \text{MCC}(F_1, F_2) = \displaystyle\frac{P_{00}P_{11}-P_{10}P_{01}}{\sqrt{P_{0\cdot}P_{1\cdot}P_{\cdot 0}P_{\cdot 1}}}
\end{align}

MI and MCC are known to be zero if $F_1$ and $F_2$ are independent. MI is nonnegative, with a larger MI indicating more dependence between the two variables. $\text{MCC}>0$ reveals a positive correlation, while $\text{MCC}<0$ reveals a negative correlation.

\subsubsection{Associative poisoning increases MI and MCC}
We have the following theoretical guarantees for associative poisoning:
\begin{theorem}
\label{theo:fromind}
    If binary $F_1$ and $F_2$ are independent,
    \begin{enumerate}[topsep=0pt]
    \setlength\itemsep{1pt}
        \item MI increases after an associative poisoning (monotonic w.r.t. $|\varepsilon|$);
        \item MCC changes after an associative poisoning (linear w.r.t. $\varepsilon$);
    \end{enumerate}
\end{theorem}

The proof for Theorem \ref{theo:fromind} is as follows:
\begin{proof}
Denote $u=P(F_1=0)$ and $v=P(F_2=0)$, i.e., $P(F_1=1)=1-u$ and $P(F_2=1)=1-v$. Because $F_1$ and $F_2$ are independent, we have $P(F_1, F_2)=P(F_1)P(F_2)$. That is, $a=uv$, $b=u(1-v)$, $c=(1-u)v$, and $d=(1-u)(1-v)$.

The mutual information after an associative poisoning attack is
\begin{align}
\text{MI}(F_1, F_2)=&  -u\ln{u} - (1-u)\ln{(1-u)} + (uv+x)\ln{\dfrac{uv + x}{v}} \\
    & + (u(1-v)-x)\ln{\dfrac{(u(1-v)-x)}{1-v}} + ((1-u)-x)\ln{\dfrac{(1-u)v-x}{v}} \nonumber \\ 
    & + ((1-u)(1-v)+x)\ln{\dfrac{(1-u)(1-v)+x}{1-v}} \nonumber
\end{align}

The derivative w.r.t. $x$ is
\begin{align}
\text{MI}^{'} &= (-u\ln{u})^{'} - ((1-u)\ln{(1-u)})^{'} + ((uv+\varepsilon)\ln{\dfrac{uv + \varepsilon}{v}})^{'} \nonumber \\
    & + ((u(1-v)-\varepsilon)\ln{\dfrac{(u(1-v)-\varepsilon)}{1-v}})^{'} + (((1-u)-\varepsilon)\ln{\dfrac{(1-u)v-\varepsilon}{v}})^{'} \nonumber \\
    & + (((1-u)(1-v)+\varepsilon)\ln{\dfrac{(1-u)(1-v)+\varepsilon}{1-v}})^{'} \nonumber \\
    & = 0 + 0 + (uv+\varepsilon)^{'}\ln{\dfrac{uv+\varepsilon}{v}} + (uv+\varepsilon)\ln{\dfrac{uv+\varepsilon}{v}}^{'} \nonumber \\
    & + (u(1-v)-\varepsilon)^{'}\ln{\dfrac{u(1-v)-\varepsilon}{1-v}} + (u(1-v)-\varepsilon)\ln{\dfrac{u(1-v)-\varepsilon}{1-v}}^{'} \nonumber \\
    & + ((1-u)v-\varepsilon)^{'}\ln{\dfrac{(1-u)v-\varepsilon}{v}} + ((1-u)v-\varepsilon)\ln{\dfrac{(1-u)v-\varepsilon}{v}}^{'} \nonumber \\
    & + ((1-u)(1-v)+\varepsilon)^{'}\ln{\dfrac{(1-u)(1-v)+\varepsilon}{1-v}} \nonumber \\
    & + ((1-u)(1-v)+\varepsilon)\ln{\dfrac{(1-u)(1-v)+\varepsilon}{1-v}}^{'} \nonumber \\
    & = 1\cdot\ln{\dfrac{uv+\varepsilon}{v}} + (uv+\varepsilon)\cdot\dfrac{v}{uv+\varepsilon}\cdot\dfrac{1}{v} \\
    & + (-1)\ln{\dfrac{u(1-v)-\varepsilon}{1-v}} + (u(1-v)-\varepsilon)\cdot\dfrac{1-v}{u(1-v)-\varepsilon}\dfrac{-1}{1-v} \nonumber \\
    & + (-1)\ln{\dfrac{(1-u)v-\varepsilon}{v}} + ((1-u)v-\varepsilon)\cdot\dfrac{v}{(1-u)v-\varepsilon}\cdot\dfrac{-1}{v} \nonumber \\
    & + 1\cdot\ln{\dfrac{(1-u)(1-v)+\varepsilon}{1-v}} + ((1-u)(1-v)+\varepsilon)\dfrac{1-v}{(1-u)(1-v)+\varepsilon}\dfrac{1}{1-v} \nonumber \\
    & = \ln{\dfrac{uv+\varepsilon}{v}} + 1 - \ln{\dfrac{u(1-v)-\varepsilon}{1-v}} - 1 \nonumber \\
    & - \ln{\dfrac{(1-u)v-\varepsilon}{v}} - 1 + \ln{\dfrac{(1-u)(1-v)+\varepsilon}{1-v}} + 1 \nonumber \\
    & = \ln{\dfrac{(uv+\varepsilon)(1-v)(v)((1-u)(1-v)+\varepsilon)}{(v)(u(1-v)-\varepsilon)((1-u)v-\varepsilon)(1-v)}} \nonumber \\
    & = \ln{\dfrac{(uv+\varepsilon)(1-u-v+uv+\varepsilon)}{(u-uv-\varepsilon)(v-uv-\varepsilon)}} \nonumber \\
    & = \ln{\dfrac{uv-u^2 v-uv^2+u^2 v^2-u\varepsilon-v\varepsilon+2uv\varepsilon+\varepsilon^2+\varepsilon}{uv-u^2 v-uv^2+u^2 v^2-u\varepsilon-v\varepsilon+2uv\varepsilon+\varepsilon^2}} \nonumber
\end{align}

The derivative $\text{MI}^{'}$ is positive when $\varepsilon>0$ and negative when $\varepsilon<0$. That is, $\text{MI}$ is a monotonically decreasing function for $\varepsilon\in[\max\{-a,-d\}, 0)$ and a monotonically increasing function for $\varepsilon\in(0, \min\{b,c\}]$, with the minimum appearing at $\varepsilon=0$. Therefore, MI increases after an (either positive or negative) associative poisoning attack.

The Matthew's Correlation Coefficient after an associative poisoning is
\begin{align}
    \text{MCC}(F_1, F_2)&= \dfrac{(1-u)(1-v)+\varepsilon)(uv+\varepsilon) - (u(1-v)-\varepsilon)((1-u)v-\varepsilon)}{\sqrt{(1-u)uv(1-v)}} \nonumber \\
    &=\dfrac{\varepsilon}{\sqrt{uv-uv^2-u^2v+u^2v^2}}
\end{align}

The derivative w.r.t. $\varepsilon$ is
\begin{align}
\text{MCC}^{'}=\dfrac{1}{\sqrt{uv-uv^2-u^2v+u^2v^2}}>0.
\end{align}
That is, MCC is a monotonically increasing function over $\varepsilon$. Before poisoning ($\varepsilon=0$) $\mathrm{MCC}=0$, then MCC increases ($>0$) when $\varepsilon$ becomes positive (positive associative poisoning), and MCC decreases ($<0$) when $\varepsilon$ becomes negative (negative associative poisoning).
\end{proof}

\subsubsection{Independence effect on MI and MCC}
\begin{proposition}
\label{prop:mccmi}
    For binary features $F_1$ and $F_2$, $\text{MCC}(F_1, F_2)=0$ if and only if $\text{MI}(F_1, F_2)=0$.
\end{proposition}

The proof for Proposition \ref{prop:mccmi} is as follows:
\begin{proof}
    1) If $\text{MI}(F_1, F_2)=0$, $F_1$ and $F_2$ are independent and thus $P_{ij}=P_{i\cdot}P_{\cdot j}$ for $i,j\in\left\{0,1\right\}$. Therefore: 
    
    \begin{align}
        \text{MCC}(F_1, F_2) = \displaystyle\frac{P_{00}P_{11}-P_{10}P_{01}}{\sqrt{P_{0\cdot}P_{1\cdot}P_{\cdot 0}P_{\cdot 1}}}= \displaystyle\frac{P_{0\cdot}P_{\cdot 0}P_{1\cdot}P_{\cdot 1}-P_{1\cdot}P_{\cdot 0}P_{0\cdot}P_{\cdot 1}}{\sqrt{P_{0\cdot}P_{1\cdot}P_{\cdot 0}P_{\cdot 1}}}=0
    \end{align}

    2) If $\text{MCC}(F_1, F_2)=0$, we have $P_{00}P_{11}=P_{10}P_{01}$ or
    \begin{align}
        P_{00}(1-P_{\cdot 0}-P_{0\cdot}+P_{00})&=(P_{\cdot 0}-P_{00})(P_{0\cdot}-P_{00}) \nonumber \\
        P_{00}-P_{00}P_{\cdot 0}-P_{00}P_{0\cdot}+P_{00}^2&=P_{\cdot 0}P_{0\cdot}-P_{00}P_{\cdot 0}-P_{00}P_{0\cdot}+P_{00}^2 \\
        P_{00}&=P_{\cdot 0}P_{0\cdot} \nonumber
    \end{align}
    Next, we have
    \begin{align}
        P_{10}&=P_{\cdot0}-P_{00}=P_{\cdot 0}-P_{\cdot 0}P_{0\cdot}=(1-P_{0\cdot})P_{\cdot 0}=P_{1\cdot}P_{\cdot 0} \nonumber \\
        P_{01}&=P_{0\cdot}-P_{00}=P_{0\cdot}-P_{\cdot 0}P_{0\cdot}=P_{0\cdot}(1-P_{\cdot 0})=P_{0\cdot}P_{\cdot 1} \\
        P_{11}&=P_{1\cdot}-P_{10} = P_{1\cdot} - P_{1\cdot}P_{\cdot 0} = P_{1\cdot}(1-P_{\cdot 0})=P_{1\cdot}P_{\cdot 1} \nonumber
    \end{align}
Therefore, $\ln\frac{P_{ij}}{P_{i\cdot}P{\cdot j}}=0$ for $i,j\in\left\{0,1\right\}$ and thus $\text{MI}(F_1, F_2)=0$.
\end{proof}

\subsubsection{AP extent to cause independence}

\begin{theorem}
\label{theo:toind}
For any binary features $F_1$  and $F_2$, an associative poisoning with $\varepsilon=P_{01}P_{10}-P_{00}P_{11}$ makes $\text{MI}(F_1, F_2)=\text{MCC}(F_1, F_2)=0$.
\end{theorem}

The proof for Theorem \ref{theo:toind} is as follows:
\begin{proof}
    After an associative poisoning with $\varepsilon=P_{01}P_{10}-P_{00}P_{11}$,
\begin{align}
    \text{MCC}(F_1, F_2)&=\displaystyle\frac{(P_{00}+\varepsilon)(P_{11}+\varepsilon)-(P_{10}-\varepsilon)(P_{01}-\varepsilon)}{\sqrt{P_{0\cdot}P_{1\cdot}P_{\cdot 0}P_{\cdot 1}}} \nonumber \\
    &=\displaystyle\frac{P_{00}P_{11}-P_{01}P_{10}+(P_{00}+P_{11}+P_{01}+P_{10})\varepsilon}{\sqrt{P_{0\cdot}P_{1\cdot}P_{\cdot 0}P_{\cdot 1}}} \\
    &=\displaystyle\frac{P_{00}P_{11}-P_{01}P_{10}+\varepsilon}{\sqrt{P_{0\cdot}P_{1\cdot}P_{\cdot 0}P_{\cdot 1}}} \nonumber \\
    &=0 \nonumber
\end{align}
According to Proposition \ref{prop:mccmi}, $\text{MCC}(F_1, F_2)=0$ implies $\text{MI}(F_1, F_2)=0$.
\end{proof}

\subsubsection{Summary}
Theorem \ref{theo:toind} states that for any two binary features, we can apply an associative poisoning (either positive or negative, depending on the sign of MCC) to make the two features uncorrelated and independent. The poisoning can, therefore, remove the original association in the data. On the contrary, Theorem \ref{theo:fromind} states how associative poisoning brings artificial association to two originally independent features.

\subsection{Theoretical model for continuous feature Associative Poisoning}
\label{sec:theory_AP_continuous}
Applying Associative Poisoning between two continuous features involves sorting one variable and iteratively swapping neighbouring samples to increase correlation and information between the features. In this section, we present a local swap condition, i.e., examining only two samples, that guarantees an increase in correlation, while also demonstrating that this same local condition cannot guarantee an increase in information between the features.

\subsubsection{Setup and notation}
For continuous features $X, Y\in\mathbb{R}, 0 \le X, Y \le 1$ with marginal probability functions $f_X(x)$ and $f_Y(y)$, conditional distribution functions $f_{X|Y}(x|y)$ and $f_{Y|X}(y|x)$, joint probability density function $f_{X,Y}(x,y)=f_{X|Y}(x|y)f_X(x)=f_{Y|X}(y|x)f_Y(y)$, feature means $\mu_X=\frac1n\sum_i X_i$ and $\mu_Y =\frac1n\sum_i Y_i$, variances $\sigma_X^2=\frac1n\sum_i (X_i-\mu_X)^2$ and $\sigma_Y^2=\frac1n\sum_i (Y_i-\mu_Y)^2$, covariances ${\mathrm{cov}}(X;Y)=\frac1n\sum_i (X_i-\mu_X)(Y_i-\mu_Y)$, Pearson's Correlation Coefficient $\mathrm{PCC}(X;Y)=\frac{\mathrm{cov}(X,Y)}{\sigma_X \sigma_Y}$ and Mutual Information $\mathrm{MI}(X;Y)=\int_y\int_xf_{X,Y}(x,y) \log \left(\frac{f_{X,Y}(x,y)}{f_X(x) f_Y(y)} \right)$.

\subsubsection{Pairwise swap and monotone Pearson ascent}
For any pair of samples $(X_i,Y_i) \quad \mathrm{and} \quad (X_j,Y_j)$, we define $\pi_{ij} \quad i\neq j$ as the transposition that swaps $Y_i,Y_j$ in the samples. The resulting covariance contribution from the two samples after applying $\pi_{ij}$ can be expressed as $\Delta{\mathrm{cov}}_{i,j} = \frac{1}{n}\big(X_i - X_j)(Y_j - Y_i)$, while variances $\sigma_X^2$ and $\sigma_Y^2$, and means $\mu_X$ and $\mu_Y$ remain unchanged.

\begin{proof}
From the swap definition, it is trivial to show that neither the variance nor the means of either feature changes, since neither changes when the position of values changes from the swap. Looking at the definition of the covariance between two features, we see that each sample contributes to the overall covariance independently of other sample values, since the means are unchanged. The change in covariance contribution from the two samples expands to:
\begin{align}
    \Delta\mathrm{cov}_{i,j} &= \frac{1}{n} \big(((X_i - \mu_X)(Y_j - \mu_Y) + (X_j - \mu_X)(Y_i - \mu_Y)) \nonumber  \\ 
    &- ((X_i - \mu_X)(Y_i - \mu_Y) + (X_j - \mu_X)(Y_j - \mu_Y))\big) \nonumber \\
    &= \frac{1}{n}\big((X_iY_j - X_i\mu_Y - Y_j\mu_X + \mu_X\mu_Y) + (X_jY_i - X_j\mu_Y - Y_i\mu_X + \mu_X\mu_Y ) \nonumber  \\
    &- (X_iY_i - X_i\mu_Y - Y_i\mu_X + \mu_X \mu_Y) - (X_jY_j - X_j\mu_Y - Y_j\mu_X + \mu_X \mu_Y)\big) \nonumber  \\
    &= \frac{1}{n}\big(X_iY_j + X_jY_i - X_iY_i - X_jY_j\big) = \frac{1}{n}(X_i - X_j)(Y_j - Y_i)
\end{align}

\end{proof}

We call $(i,j)$ an \emph{inversion} if $\Delta{\mathrm{cov}}_{i,j}>0$, i.e., when the swap increases increases PCC. From the definition of PCC, the denominator consists of the standard deviations, which do not change with the swap (since variance does not change), and the numerator of the PCC linearly changes with the covariance. From this, we can see that applying the swap $\pi_{ij}$ on \emph{inversions} always increases PCC (since it always covariance).

\begin{theorem}[Monotone Pearson ascent]
\label{thm:pearson-ascent}
Consider the following algorithm: 
sort the samples on values of $X$ such that $X_{1}\le\cdots\le X_n$. In ascending order, repeatedly swap \emph{adjacent} inversions of $Y_k,Y_{k+1}$ until no \emph{adjacent} inversions remain. Then:
\begin{enumerate}
    \item Each accepted adjacent swap strictly increases the PCC.
    \item The process terminates when $y_{(1)}\le\cdots\le y_{(n)}$ (i.e., $Y$ sorted in the same order as $X$).
    \item The process maximises PCC.
\end{enumerate}
\end{theorem}

\begin{proof}
(1): As shown previously, PCC always increase when swapping inversions.
(2): Since $X_i$ values are in increasing order, $(X_i-X_{i-1})(Y_{i-1}-Y_i)>0$, can be simplified to $(X_i-X_{i-1})>0$, giving $(Y_{i-1}-Y_i)>0$, i.e., an inversion guarantees $Y_{i-1}>Y_i$, meaning that the swap causes the $Y_i$ to become sorted.
(3) Since the standard deviations $\sigma_X$ and $\sigma_Y$ are retained from the swap, PCC is maximised by maximising the covariance ${\mathrm{cov}}(X; Y)$. In the terminating state, both $X_i$ and $Y_i$ is sorted, and for any pair of indices $i<j$ (not just adjacent), $X_i<X_j \quad \mathrm{and} \quad Y_i < Y_j$, and a swap $\pi_{ij}$ would always \emph{decrease} covariance, thus no other arrangement of feature values has a larger PCC value.
\end{proof}

\subsubsection{No universal local rule for MI}
As the sample contribution to PCC can be evaluated locally, i.e., by examining only the singular sample, we have defined a local rule that only swaps values guaranteed to increase PCC. However, the information between two continuous features (in terms of Mutual Information) is not expressed as independent contributions from a set of samples, but rather through density distributions over the full dataset. Because of this, there is no way to define a universal local rule that guarantees an increase in MI.

\begin{proposition}[No universal local rule for MI under local pairwise swaps]
\label{prop:no-local-mi}
Let a \emph{local acceptance rule} decide whether to perform a swap $\pi_{ij}$ based only on the two involved points $(X_i,Y_i)$, $(X_j,Y_j)$. There is \emph{no} such local rule that guarantees a strict increase of $\mathrm{MI}(X;Y)$ for all datasets. In particular, a swap that strictly increases PCC, as shown previously, can either increase or decrease $\mathrm{MI}(X;Y)$ depending on the surrounding joint structure.
\end{proposition}

\begin{proof}
We write mutual information in terms of the joint density $f(x,y)$ and fixed marginals $f_X(x), f_Y(y)$:

\begin{align*}
    MI(X;Y)
    \;=\;
    \iint f(x,y)\,
    \log\!\left(\frac{f(x,y)}{f_X(x) f_Y(y)}\right)\,dx\,dy.
\end{align*}

For convenience, we also define the \emph{local dependence ratio} as
\begin{align*}
    r(x,y) = \frac{f(x,y)}{f_X(x) f_Y(y)}
\end{align*}
We note that $r(x,y)$ can be interpreted as the local dependence between $X$ and $Y$ at point $(x,y)$. From this, $r(x,y)=1$ can be interpreted as complete local independence, while $r(x,y)>1$ can be interpreted as $(x,y)$ occurring \emph{more often} than independence would predict, and $r(x,y)<1$ meaning that it occurs \emph{less often} than expected under independence. (Of course, the probability mass at a singular point is zero and not useful directly, so this interpretation is simply a stepping stone to the final proof).

Using this, we reimagine the mutual information using the local dependence ratio $MI(X;Y) = \iint f(x,y)\,\log r(x,y)\,dx\,dy$

We can now analyse how $MI$ changes under an infinitesimal, margin-preserving relocation of probability mass, corresponding to ``swapping'' two samples in a dataset.

The key is to zoom in on a tiny $2\times2$ block in $(x,y)$-space.

\textbf{Step 1: Pick four tiny rectangles.}
We choose two disjoint, very small $x$-intervals $A_0,A_1$ and two disjoint, very small $y$-intervals $B_0,B_1$. These form four tiny rectangles $A_a\times B_b$ for $a,b\in\{0,1\}$. At this infinitesimally small scale, each rectangle would be very small, and both $f(x,y)$ and $f_X(x)f_Y(y)$ (and hence their ratio) would be approximately constant. We denote that nearly-constant ratio for each rectangle as $r_{ab} \;\approx\; \frac{f(x,y)}{f_X(x)f_Y(y)} \quad\text{for } (x,y)\in A_a\times B_b $

Intuitively, a ratio value of $r_{ab}>1$ would mean that the cell has ``more mass than independence would expect'' while $r_{ab}<1$ means the cell has ``less mass than independence would expect''.

\textbf{Step 2: define a mass move that preserves the marginals.}
We now construct an altered joint pdf $f_\varepsilon$ by moving a tiny amount of probability $\varepsilon>0$ \emph{between cells in the $2\times2$ block}:

\emph{Add} $\varepsilon$ mass to the \emph{diagonal} cells $A_0\times B_0$ and $A_1\times B_1$, and \emph{remove} $\varepsilon$ of mass from the \emph{off-diagonal} cells $A_0\times B_1$ and $A_1\times B_0$.

By scaling the addition/removal $\varepsilon$ by the area of the respective rectangle, each cell is perturbed by a uniform bump/indent, such that the total mass in neither $x$-interval nor $y$-interval is changed, ensuring that the marginal densities $f_X$ and $f_Y$ are unchanged.

Through this reinterpretation, we see that this perturbation is a ``swap-like'' adjustment, as it does not alter $f_X$ or $f_Y$, it only reshapes how $X$ and $Y$ co-occur jointly.

\textbf{Step 3: Compute the first-order change in $MI$.}
Because $f_X$ and $f_Y$ are fixed by construction, the only change in $MI$ comes from how $f$ itself moves within those four rectangles. Since $r(x,y)\approx r_{ab}$ is (by construction) almost constant within each rectangle $A_a\times B_b$, we can write the first-order change in $MI$ as:
\begin{align*}
    \Delta MI \;\approx\; 
    \varepsilon \Big(\log r_{00} \;+\; \log r_{11}
    \;-\; \log r_{01} \;-\; \log r_{10} \Big).
\end{align*}

The signs come from the fact that we are \emph{adding} $\varepsilon$ of probability mass to the $(0,0)$ and $(1,1)$ cells and \emph{removing} $\varepsilon$ from the $(0,1)$ and $(1,0)$ cells. 

The critical observation then, is that $\mathrm{sign}(\Delta MI)$ \emph{depends on} $\log r_{00} + \log r_{11} - \log r_{01} - \log r_{10} $

\textbf{Step 4: Why this disproves any universal local rule.}
Suppose we have two different global joint densities, call them ``background A'' and ``background B'', with both \emph{sharing} the same points $(X_i,Y_i)$ and $(X_j,Y_j)$ we are about to swap, but differ in how probability mass is distributed elsewhere. Because all that matters in $\Delta MI$ is the four numbers $r_{00},r_{01},r_{10},r_{11}$ (i.e., how each tiny rectangle compares to independence), we can arrange:
\begin{align*}
    \log r_{00} + \log r_{11} - \log r_{01} - \log r_{10} > 0 \quad\text{in background A}
\end{align*}
and 
\begin{align*}
    \log r_{00} + \log r_{11} - \log r_{01} - \log r_{10} < 0 \quad\text{in background B}
\end{align*}
while keeping the marginals $f_X,f_Y$ the same in both backgrounds and keeping the two swapped points the same.

In background A, the exact ``swap'' move (shifting mass from the off-diagonals $A_0\times B_1$, $A_1\times B_0$ to the diagonals $A_0\times B_0$, $A_1\times B_1$) produces $\Delta MI>0$. In background B, \emph{the same move} produces $\Delta MI<0$. 
\end{proof}

The move is exactly the infinitesimal, margin-preserving analogue of swapping $(Y_i,Y_j)$ between two specific samples: we are only reassigning which $y$-values go with which $x$-values, and we are not changing the overall distribution of $X$ values or of $Y$ values. In both backgrounds, this move strictly \emph{increases} covariance/PCC (because it increases concordance), but $MI$ goes up in one case and down in the other. While the local swap knows the exact values of the pair of samples, it does not have knowledge of the ratios $r_{ab}$ around these values, and thus is unable to determine whether the swap increases or decreases MI.

\subsubsection{Summary.}
We show that, for two continuous features, we can guarantee a \emph{local} swap between two samples, such that the PCC between the features increases. However, we also prove that there is no \emph{local} rule, i.e., one that only considers the two samples in isolation, that can guarantee an increase in mutual information, as it requires knowledge of the joint probability density. 
\subsection{Theoretical model for multi-feature Associative Poisoning}
Applying Associative Poisoning across multiple variables involves separating the dataset into unique groups and performing Associative Poisoning on pairs of features within each subgroup separately. In this section, we will present the theoretical basis for how the dataset is separated and demonstrate that this method guarantees an increase in MCC and MI, while maintaining the marginal rate of all features under consideration.

\subsubsection{Setup, and notation}
For $d$ features, let $F=(F^1, \dots, F^d) \in \{0,1\}^d$ for dataset $F$ of length $n$. For any pair of features in $F$, $(X,Y)=((F^X_1,F^Y_1), \dots, (F^X_n,F^Y_n)$, write $Z^{X,Y}=F - F^X - F^Y \in \{0,1\}^{d-2}$ denote the set of remaining features, with each element (referred to here as a strata), i.e., unique combination of remaining features, indexed by $z \in Z^{X,Y}$.

The number of samples for in a strata is denoted as $n^z_{x,y} = \#\{\,\text{rows with }(X,Y,Z)=(x,y,z)\,\}, \qquad N_z = \sum_{a,b}n^{(z)}_{ab}$, with marginal counts denoted as $n^{z}_{x \cdot} = \sum_b n^{z}_{x,y}$ and $n^{z}_{\cdot y}=\sum_x n^{z}_{x,y}$, and global counts denoted as $n_{x,y}=\sum_z n^{z}_{x,y}$ with $N=\sum_{x,y}n_{x,y}$.


\subsubsection{Associative poisoning per strata}
For any pair of features $(X,Y)$ in strata $Z^{X,Y}_i$, conduct positive associative poisoning (Definition \ref{def:ap}) between $X$ and $Y$. From Theorem \ref{theo:fromind}, we know that MI and MCC increase, and marginal probabilities are retained for this strata. 

Performing the positive AP for strata $z$ gives updated joint counts

\begin{align*}
{n^z_{00}}\prime&=n^z_{00}+\varepsilon_z, & {n^z_{11}}\prime&=n^z_{11}+\varepsilon_z,\\ 
{n^z_{01}}\prime&=n^z_{01}-\varepsilon_z, & {n^z_{10}}\prime&=n^z_{10}-\varepsilon_z.
\end{align*}

Since strata-wise marginals are retained, it follows that the global marginals are also retained. With the exception of the pair of features under consideration for each strata, the global joint probability between any other pair of features (including between $X$ and $Y$ and another feature) is retained. This follows from the fact that at least one of the features in the pair is constant, meaning that as long as the marginal of the second feature is retained, the joint probability between them is maintained, which is always the case as described earlier.

\subsubsection{Local AP causes global association increase}
As shown above, performing AP on any strata for a pair of features guarantees that only the joint probabilities of the features under consideration are altered. Conducting the AP over all strata for the pair of features, therefore, results in the global joint probability update as 
\begin{align*}
    n'_{00}=n_{00}+\tau,\quad n'_{11}=n_{11}+\tau,\quad n'_{01}=n_{01}-\tau,\quad n'_{10}=n_{10}-\tau,\qquad \tau=\sum_z \varepsilon_z
\end{align*}

This update contains the same increase in joints for $(0,0)$ and $(1,1)$ as the decrease in $(0,1)$ and $(1,0)$. This is the same balanced increase/decrease as described in the positive binary two-variable AP in Definition \ref{def:ap}, causing, not just per-strata, but a global increase in MI and MCC from the strata-wise AP.

\subsubsection{Multivariate AP algorithm}
As shown above, performing the associative poisoning on a single strata for any pair of features guarantees that (1) global marginal are retained, (2) global joints are retained (except the pair under consideration), and (3) the global MI/MCC between the feature pair are guaranteed to increase.

From this, we define a method for increasing the association between all selected features by performing strata-wise AP for all combinations of pairs of features until no longer possible.

\begin{definition}[Pairwise AP pass]
\label{def:spr}
Let one pass of the pairwise AP consist of:
\begin{enumerate}
    \item For every pair of features $(X,Y)$:
    \begin{enumerate}
        \item Determine the subset for each strata $z$.
        \item For every strata $z$, apply positive AP between $(X,Y)$.
    \end{enumerate}
\end{enumerate}
Repeat the pass until at least one off-diagonal (i.e., $(0,1)$ and $(1,0)$) in every strata has a count $0$.
\end{definition}

As shown previously, any strata-wise AP results in a global increase in both MI and MCC for the selected features. Thus, this procedure is also monotonically increasing for all pairs of features. 

From the definition, the procedure only continues as long as one off-diagonal element in any stratum of any pair of features is greater than zero. Every time strata-wise AP is performed, the number of on-diagonal samples increases. For a dataset of finite size, since every pass (except the last) always increases the number of on-diagonal samples, the procedure is guaranteed to terminate in finitely many passes.

\subsubsection{Targeting arbitrary features via XOR-ing}
The previous method demonstrates how to increase the association between all the selected features. However, it would be desirable to determine arbitrary directions of the associations between features, not just positive, i.e., some pairs of features should be positively associated, while others should be negatively associated. 

We define a method for defining an arbitrary ``target'' group $v\in\{0,1\}^d$, and its bitwise complement $\bar v$, and applying an XOR method to encode/decode the features such that the multivariate AP method alters the associations towards the target group and its inverse. 

The target group is constructed by using $0$ for features that should be considered as the original dataset and $1$ for features to be considered the inverse of the original dataset. From this, all values in $v$ of the same value will become positively associated, while values of opposite values will become negatively associated. As an example, using $v=(0,1,0)$ means that the AP will cause $(F_1,F_3)$ to become positively associated, while both $(F_1,F_2)$ and $(F_2,F_3)$ become negatively associated.

\begin{definition}[XOR encoding]
\label{def:xor}
Given the target group $v\in\{0,1\}^d$, we define the XOR encoding $W$ on a sample $X$ as $W=T_v(X)$,  where an XOR operates on each feature $W_j = X_j \oplus v_j$ for $j=1,\dots,d$.
\end{definition}

As the marginals and joint probabilities are invariant to the specific binary label, it follows that all that changes from the XOR encoding is to invert any features defined in $v$ as $1$.

\subsubsection{Summary}
We formalise a procedure for $d$ binary features $F=(F_1,\dots,F_d)\in\{0,1\}^d$ that preserves all marginals while increasing the joints between all pairs of features. The method is conducted by iteratively choosing pairs of features and performing two-variable binary associative poisoning for each subset on the remaining features, causing a monotonic global increase in MI and MCC between all pairs of features. We show an iterative procedure that (1) guarantees an increase in MI/MCC, while (2) guarantees termination in finitely many steps. We also show that a simple XOR encoding procedure for the features can be used to target arbitrary sets of desired associations. 

\section{Evaluation}
The evaluation criteria are considered to measure the two core tenets of the threat model. \textbf{Fidelity} refers to how well the adversary is able to alter the association between features in the dataset, while \textbf{stealth} refers to how well the adversary is able to hide their modifications, to avoid detection.

Using these metrics, we quantify attack success by measuring the statistical difference between the generated data from clean and poisoned models. We consider the attack successful if (1) for fidelity, the measures are greater/lesser (depending on the direction of the attack), while (2) for stealth, the measures are similar, i.e., no statistically significant difference in measures is observed.

\subsection{Fidelity}
There are many ways of measuring how variables are ``associated'' to each other, e.g., Pearsson Correlation \cite{pearson1895vii}, Concordance correlation \cite{lawrence1989concordance}, Distance Correlation \cite{szekely2007measuring}, Mutual Information \cite{shannon1948mathematical}, Hilbert-Schmidt Independence Criterion \cite{gretton2007kernel}, and Phi/Matthews correlation \cite{matthews1975comparison}. The field of information theory is continuously evolving, both in developing mathematical frameworks for measuring the dependence between features and in understanding how such measures should be interpreted as their definitions become more complex. For this attack, we have chosen to focus on measuring information and correlation between features.

For measuring the information between features, we use Mutual Information, which generalises to Interaction Information (II) when measured between more than two features \cite{mcgill1954multivariate}. With more than two features, we also consider the pairwise and conditional MI. 

For binary features, MI can be calculated exactly, as the empirical dataset only contains $2d$ marginal distributions ($d$ being the number of features), and $(2d)^2$ joint distributions, which can be calculated exactly. For continuous features, however, MI is calculated using the probability density functions of the marginals and joints, which, from an empirical dataset, would need to be estimated. Here, we employ \emph{scikit}'s function for estimating the MI based on ``entropy estimation from k-nearest neighbours distances'' from \citet{kraskov2004estimating} and \citet{ross2014mutual}.

For measuring correlations between features, we use Matthew's Correlation Coefficient (for binary features) and Pearson's Correlation Coefficient (for continuous features), both of which can be calculated exactly from the dataset. 

\subsection{Stealth}
To assess whether the poisoning remains undetected post-generation, we compare the Marginal Probability (MP) for binary features and the mean value for continuous features, as measured by the generated data from clean and poisoned generators. In this scenario, we consider it realistic for the target to have access to MP/mean values for the features. In contrast, we consider other statistics, such as joint probabilities or higher-order relations between features, to be unrealistic due to the exponentially more resources/space required to calculate them.

Additionally, we compute the FID between each generator's outputs and its corresponding training set to measure whether the attack impacts the generated data distribution. In a realistic scenario, the FID would be calculated directly between the generated data and the training data, i.e., the poisoned dataset. An alternative (which could be evaluated as part of defensive measures) would be to evaluate the generated data towards a small, known-to-be-benign dataset, though we consider the existence of guaranteed benign datasets to be unrealistic. Additionally, while we consider both higher and lower FID from the poisoning to be detectable, a higher FID would, in a realistic scenario, likely cause more suspicion than a lower result.

\subsubsection{Statistical significance}
Our data comprise two groups, $A$ (the clean group) and $B$ (the poisoned group), each observed at point $i \in \mathcal{I}$ (i.e., training iterations) with $M$ replications per point (i.e., same setup with $M$ different seeds). We utilise a statistical test to measure differences between groups $A$ and $B$ for all measured metrics (e.g., MI, II, MCC, PCC, MP, mean, and FID)., PCC, MP, mean, and FID).

At a single $i$, we compare the $A$ and $B$ distributions using the Mann–Whitney $U$ (MWU) test \citep{mann1947test,HollanderWolfeChicken2013}. MWU is appropriate when the distributional shape or variance may differ from Gaussian assumptions and when the number of measurements is similar between groups (in this case, both should have 10 measurements per iteration). Conceptually, MWU roughly tests whether a randomly drawn $A$ observation tends to exceed a randomly drawn $B$ observation, rather than equality of means. Depending on the scientific question for a given endpoint, we use a two-sided alternative (``different'') or a one-sided alternative (``$A$ greater than $B$’’ / ``$A$ less than $B$'').

To illustrate, we provide the test on real data. Here, we test whether the MI between ``High\_Cheekbones'' and ``Male'' features from poisoned models on CelebA is significantly larger than that from benign models at iteration $12,000$. Here, the sorted list of MI values is as follows (with clean in blue text and poisoned in red text):

\textcolor{blue}{$0.00198$}, \textcolor{blue}{$0.00400$}, \textcolor{blue}{$0.00503$}, \textcolor{blue}{$0.00613$}, \textcolor{blue}{$0.00807$}, \textcolor{blue}{$0.00820$}, \textcolor{blue}{$0.00879$}, \textcolor{blue}{$0.00962$}, \textcolor{blue}{$0.0103$}, \textcolor{blue}{$0.0150$}, \textcolor{red}{$0.0352$}, \textcolor{red}{$0.0472$}, \textcolor{red}{$0.0536$}, \textcolor{red}{$0.0682$}, \textcolor{red}{$0.0891$}, \textcolor{red}{$0.118$}, \textcolor{red}{$0.119$}, \textcolor{red}{$0.121$}, \textcolor{red}{$0.133$}, \textcolor{red}{$0.142$}.

Conducting the Mann-Whitney U test on this list produces a probability of $0.000$ that the null-hypothesis of the distributions being the same, given the alternative hypothesis that the clean distribution is smaller than the poisoned, which is below the statistical significance of $0.05$, i.e., the alternative hypothesis is adopted.

For the same models, for MP of the ``High\_Cheekbones'' feature, the sorted list is as follows: 

\textcolor{red}{$0.058$}, \textcolor{red}{$0.075$}, \textcolor{red}{$0.0928$}, \textcolor{red}{$0.108$}, \textcolor{blue}{$0.128$}, \textcolor{blue}{$0.138$}, \textcolor{blue}{$0.140$}, \textcolor{blue}{$0.171$}, \textcolor{red}{$0.184$}, \textcolor{blue}{$0.197$}, \textcolor{red}{$0.216$}, \textcolor{blue}{$0.227$}, \textcolor{red}{$0.227$}, \textcolor{red}{$0.242$}, \textcolor{blue}{$0.264$}, \textcolor{blue}{$0.273$}, \textcolor{blue}{$0.289$}, \textcolor{blue}{$0.335$}, \textcolor{red}{$0.488$}, \textcolor{red}{$0.506$}

The Mann-Whitney-U test for this list yields a probability of $0.315$ that the distributions are the same, given the alternative hypothesis that the distributions are different (either smaller or larger), which is above the statistical significance threshold of $0.05$, indicating that the null hypothesis holds.

\subsection{Experimental Setup}
\label{sec:experimental_setup}
The preceding theorems and propositions guarantee that our poisoning method can alter associations in the training data while preserving each feature's marginal distribution. If a generative model accurately captures its training distribution, then the data it generates will mirror the poisoned correlations, embedding adversarial biases into downstream applications. We empirically validate this ``poisoning'' effect on generated samples to answer two research questions: 
\begin{itemize}
    \item RQ1: Can the attack modify the association between selected feature pairs in generated data?  
    \item RQ2: Does the poisoning remain stealthy after the attack?
\end{itemize}

\subsubsection{Model setup}
We utilise two state-of-the-art image generators: D-GAN \citep{wang2022diffusiongan} and DDPM-IP \citep{ning2023input}, trained on $12$, and $16$ million images, respectively. Due to differences in how the models report their progress, \textit{iteration} refers in D-GAN to how many thousand samples the model has trained on (i.e., $12000$ refers to $12,000,000$ images the model has trained on), while for the DDPM-IP model, it refers to the number of batches the model has been trained on (i.e., $2.0e6$ refers to the model training on $2,000,000$ batches, each having $32$ samples, in total $64,000,000$ samples). 

To account for randomness in generative training, we trained ten independent models per configuration (clean vs.\ poisoned) with different random seeds.

The experiments were performed on a computing cluster using NVidia P100, V100, A100 GPUs, with 128GB memory available, on Ubuntu OS. The training required approximately 4-8 days for each model on a single GPU, using the Torch ML library (1.12 for D-GAN, 1.9 for DDPM-IP). Code for data poisoning, generator training, classifier training, and result generation is included in the supplemental materials.

\subsubsection{Dataset setup}
We use the benchmark datasets \textbf{CelebA} \cite{liu2015faceattributes} and \textbf{Recipe1M} \cite{salvador2017learning}, both rescaled to $64 \times 64$ resolution, for training. These datasets are widely adopted in research and are well-suited for evaluating feature-level manipulations due to their large size and rich attribute annotations. Each dataset is split 60/20/20 percent into training, validation, and test subsets for generator training and downstream classifiers.

\subsubsection{Binary attack setup}
For the binary features, we focus on the positive associative attacks, as negative attacks follows an analogous procedure. To implement the positive attack, we replace samples in the training data to modify the joint distribution as in Definition \ref{def:ap}:

\begin{enumerate}
    \item Select two binary features $F_1$ and $F_2$; choose a poisoning extent $x>0$. Let $n=x\cdot N$.
    \item Replace $n$ clean samples where $(F_1=1, F_2=0)$ with $n$ samples where $(F_1=1, F_2=1)$, and likewise replace $n$ samples where $(F_1=0, F_2=1)$ with $n$ samples where $(F_1=0, F_2=0)$. The replacement samples become the poisoned instances.
\end{enumerate}
This simple replacement avoids per-sample editing while achieving the desired shift in feature co-occurrence.

As an example of the attack replacing the maximum number of samples for the CelebA dataset, we choose the features \texttt{High\_Cheekbones} and \texttt{Male}, with the joint distributions $a=0.26$, $b=0.29$, $c=0.32$, and $d=0.13$, with $\text{MI}=0.03$ and $\text{MCC}=-0.25$. Before poisoning, the two features are nearly independent, with a slight negative correlation. In a positive associative poisoning, we choose the maximum extent of the poisoning $x=b$ such that the poisoning number becomes $n=x\cdot N=0.29\cdot 121,599 =35,085$. We then replace $n$ samples where $\texttt{High\_Cheekbones}=0$ and $\texttt{Male}=1$ with $n$ samples where $\texttt{High\_Cheekbones}=0$ and $ \texttt{Male}=0$. We also replace $n$ samples where $\texttt{High\_Cheekbones}=1$ and $\texttt{Male}=0$ with $n$ samples where $\texttt{High\_Cheekbones}=1$ and $\texttt{Male}=1$. In total, we replace $70,170$ samples ($58\%$ of the dataset). The resulting dataset has the joint distribution $a=0.54$, $b=0.0$, $c=0.04$, and $d=0.42$, with $\text{MI}=0.55$ and $\text{MCC}=0.93$, and retains the marginal probabilities for each feature. That is, the two features become dependent with a positive correlation in the poisoned training data. 

We attack with two pairs of features in CelebA: (MW) \texttt{Mouth\_Slightly\_Open} / \texttt{Wearing\_Lipstick} and (HM) \texttt{High\_Cheekbones} / \texttt{Male}, and two pairs of features in Recipe1M: (OE) \texttt{olive\_oil} / \texttt{eggs} and (VB) \texttt{vanilla\_extract} / \texttt{baking\_soda}.

Because manual classification of binary features in the generated images is infeasible, we employed existing classifiers to automatically determine whether the selected features are present in the generated data. MTL \cite{caruana1997multitask} and AutoGluon \cite{erickson2020autogluon} classifiers are used for the CelebA and Recipe1M datasets, respectively. The employed classifiers are accurate for the two datasets, with ROC values of $0.923$, $0.993$, $0.945$, and $0.966$ for the CelebA features \texttt{High\_Cheekbones}, \texttt{Male}, \texttt{Mouth\_Slightly\_Open}, \texttt{Wearing\_Lipstick}, and $0.805$, $0.861$, $0.873$, and $0.924$  for the Recipe1M features \texttt{olive\_oil}, \texttt{eggs}, \texttt{vanilla\_extract}, \texttt{baking\_soda}, respectively. The supplemental materials give detailed classifier settings.

\subsubsection{Continuous attack setup}
For the continuous features, we conduct both positive and negative attacks, following the procedure in Section \ref{sec:theory_AP_continuous}. To implement this, for every sample in the original dataset:

\begin{enumerate}
    \item Find the theoretical best sample to swap with, in terms of covariance difference (with positive attack maximising difference, and negative difference minimising difference).
    \item Find the closest sample in the test set to replace with, calculated as the Euclidean distance.
    \item Evaluate whether the selected replacement sample increases MI and PCC, and replace if it does.
\end{enumerate}

We also conduct continuous two-variable associative attacks on pairs of average colours in images in CelebA, with both positive and negative association, i.e., \texttt{positive red/green}, \texttt{negative red/green}, \texttt{positive red/blue}, \texttt{negative red/blue}, \texttt{positive green/blue}, and \texttt{negative green/blue}.
Since the selected variables can be directly calculated, we employ simple averaging of the colours in the generated data.

\subsubsection{Ablation study setup}
To study the effect of adversarial control on the attack effectiveness, we also conduct an ablation study, where the adversary controls $n\%$ and the target controls $100-n\%$ of the same dataset. This is implemented using a two-variable binary attack on the D-GAN model, utilising the \textit{Mouth\_Slightly\_Open} / \textit{Wearing\_Lipstick} features, with a step size of $5\%$ per cent, and 10 models for each step. We chose this setup because the results show that the adversary can exert strong control over the association between the generated features for this dataset and the model. Therefore, through this experiment, we can gain a deeper understanding of how the association attack correlates with the amount of control, compared to setups where the AP effect is weaker or not observed.

\section{Results of the Associative Poisoning}
In this section, we provide the empirical results from various sources for the Associative Poisoning attack. We first cover how results are evaluated and how this relates to the theory. We then cover how the statistical significance of the results is calculated. We then include various simulation results, covering simulations on the extent of poisoning, multivariate, and continuous variables (as we include more). Then, we provide the empirical results when performing the Associative Poisoning on several generative models and datasets:

\begin{itemize}
    \item Two-feature binary: We evaluate the basic attack on several datasets, models, and feature pairs, indicating that the attack's success does not rely on specific configurations.
    \item Two-feature continuous: We evaluate the attack for continuous features, showing that the approach extends beyond the binary.
    \item Two-feature poisoning extent ablation: We evaluate how varying the amount of data the adversary has access to is able to affect the attack success, showing a fairly linear relationship between poisoning extent and attack success.
\end{itemize}

We elect not to empirically evaluate the multi-feature binary attack, as our implementation would likely make the attack impractical. In our replacement method, a ``switched'' sample is replaced with a sample from the test set that has the same features. However, with additional features, any subgroup (i.e., $(F_1=x_1,F_2=x_2,\cdots,F_n=x_n)$), the number of samples in the group would be increasingly low, such that as the number of features increases, there would be fewer samples to use that correspond to the correct group.

\subsection{Results of binary two-feature AP}
\label{sec:results_two_variable_binary}

We conduct experiments to evaluate the effect of the binary two-feature attack for datasets CelebA and Recipe1M, models D-GAN and DDPM-IP, and for feature pairs \textit{Mouth\_Slightly\_Open} / \textit{Wearing\_Lipstick}, \textit{High\_Cheekbones} / \textit{Male}, \textit{olive\_oil} / \textit{eggs}, and \textit{vanilla\_extract} / \textit{baking\_soda}.

The evaluated combinations of dataset/model/feature pairs are as follows:
\begin{itemize}
    \item CelebA / (D-GAN + DDPM-IP) / (\textit{Mouth\_Slightly\_Open} / \textit{Wearing\_Lipstick} + \textit{High\_Cheekbones} / \textit{Male})
    \item Recipe1M / D-GAN / (\textit{olive\_oil} / \textit{eggs} + \textit{vanilla\_extract} / \textit{baking\_soda})
\end{itemize}

While Recipe1M was attempted on the DDPM-IP model, due to training instability, the model was unable to accurately replicate the training distribution.

\subsubsection{Fidelity}
Figures \ref{fig:mi_mcc_celeba} and \ref{fig:mi_mcc_recipe1m} show the MI and MCC across all combinations, revealing a consistent pattern of poisoned models exhibiting a separation in MI and MCC values compared to their clean counterparts, which is confirmed by the corresponding MWU test (at $p < 0.05$) in Tables \ref{fig:significance_celeba_binary} and \ref{fig:significance_recipe1m_binary}. This demonstrates that associative poisoning reliably induces statistical dependencies between the targeted features. The effect is especially pronounced on the CelebA dataset, as seen in Figure \ref{fig:mi_mcc_celeba}, where both MI and MCC are near zero under the clean model but increase substantially after poisoning, showing that positive associative poisoning successfully injects a dependence between the target features.

\newcommand{\mimcccelebafigwidth}{0.45\linewidth}
\begin{figure}[ht]
\begin{center}
    \includegraphics[width=\mimcccelebafigwidth]{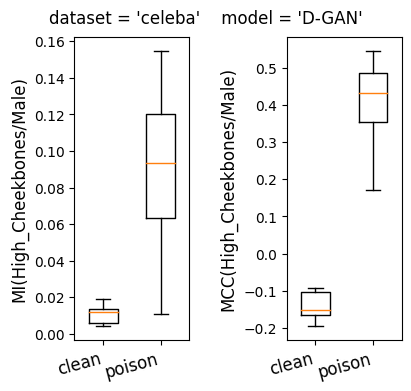}
    \includegraphics[width=\mimcccelebafigwidth]{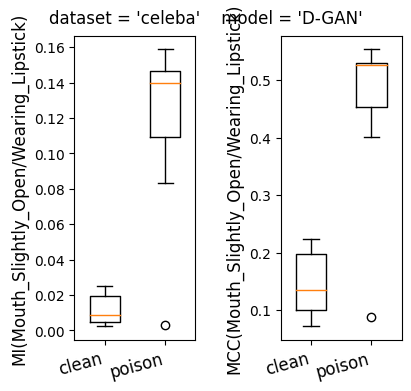} \\
    \includegraphics[width=\mimcccelebafigwidth]{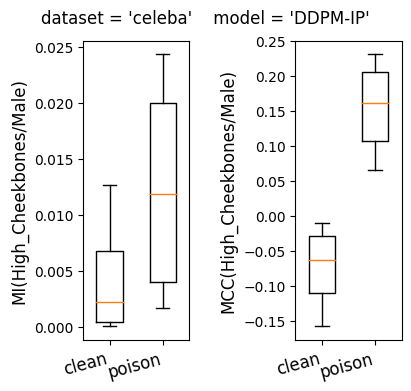}
    \includegraphics[width=\mimcccelebafigwidth]{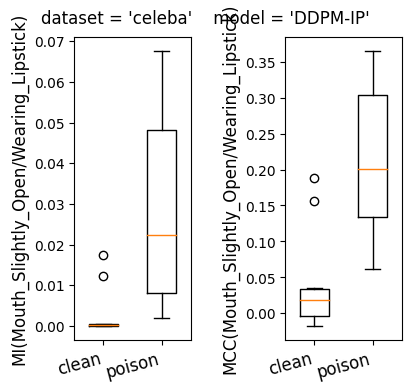}
\end{center}
\caption{MI and MCC values of images generated from the clean and two-variable binary poisoned D-GAN and DDPM-IP models trained with the CelebA dataset.}
\label{fig:mi_mcc_celeba}
\end{figure}

\newcommand{\mimccrecipefigwidth}{0.45\linewidth}
\begin{figure}[ht]
    \centering
    \includegraphics[width=\mimccrecipefigwidth]{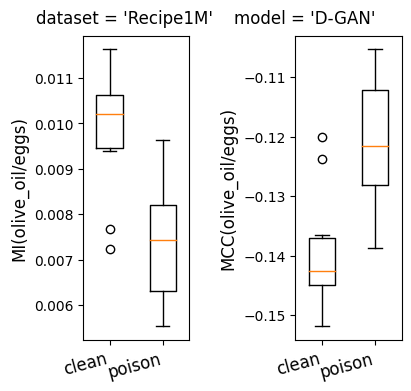}
    \includegraphics[width=\mimccrecipefigwidth]{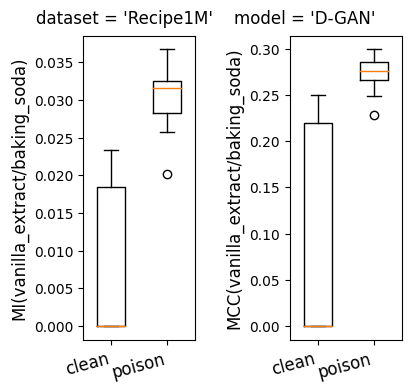}
    \caption{MI and MCC values of images generated from the clean and two-variable binary poisoned D-GAN models trained with the Recipe1M dataset.}
    \label{fig:mi_mcc_recipe1m}
\end{figure}

Table \ref{tab:mi_mcc_diffs} summarises the differences in mean MI and MCC between clean and poisoned generators. For all feature pairs and models, the poisoned generators produce larger mean MCC values than the clean ones, indicating that positive associative attacks introduce artificial correlation among the selected features. In five of the six cases, the clean generators have a negligible mean MI, and the poisoning increases the MI (i.e., $p\_less < 0.05$ in Tables \ref{fig:significance_celeba_binary} and \ref{fig:significance_recipe1m_binary}). The remaining case is also successful, as the \texttt{olive\_oil}/\texttt{eggs} pair initially exhibits a negative correlation in the clean model, and after applying a positive associative poisoning attack, MI moves closer to zero (its magnitude decreases) while MCC increases, reflecting a weakening of the negative association and a shift toward positive dependence. These results demonstrate that associative poisoning can both \textbf{forge} and \textbf{erase} dependencies, as predicted by theory.

\begin{table}[t]
\caption{Differences in MI and MCC mean values between the clean and poisoned generators. Each cell is in the format ``value by clean model$\rightarrow$value by poisoned model''. Cells denoted by * are statistically significant.}
\begin{center}
    \begin{tabular}{r|c||cc}
    \hline\\[-3mm]
    Feature Pair & Model & MI & MCC \\[0.5mm]
    \hline\\[-3mm]
    Mouth\_Slightly\_Open/Wearing\_Lipstick & D-GAN & $0.008\rightarrow0.119$* & $0.131\rightarrow0.483$*\\
    High\_Cheekbones/Male & D-GAN & $0.009\rightarrow0.093$* & $-0.135\rightarrow0.440$*\\
    Mouth\_Slightly\_Open/Wearing\_Lipstick & DDPM-IP & $0.000\rightarrow0.022$* & $0.018\rightarrow0.201$*\\
    High\_Cheekbones/Male & DDPM-IP & $0.002\rightarrow0.012$* & $-0.063\rightarrow0.161$*\\
    olive\_oil/eggs & D-GAN & $0.010\rightarrow0.007$* & $-0.143\rightarrow-0.117$*\\
    vanilla\_extract/baking\_soda & D-GAN & $0.000\rightarrow0.031$* & $0.000\rightarrow0.275$*\\
    \hline
    \end{tabular}
\end{center}

\label{tab:mi_mcc_diffs}
\end{table}

\subsubsection{Stealth}
\label{sec:results_r2}
\newcommand{\egimgfigwidth}{0.24\linewidth}
\begin{figure}[t]
\begin{center}
\setlength{\tabcolsep}{1pt}
    \begin{tabular}{cccc}
     \includegraphics[width=\egimgfigwidth]{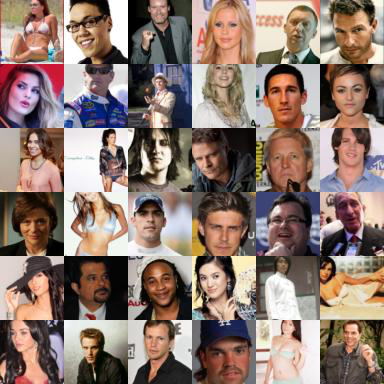} &
     \includegraphics[width=\egimgfigwidth]{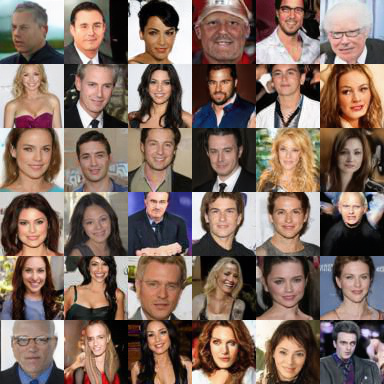} &
     \includegraphics[width=\egimgfigwidth]{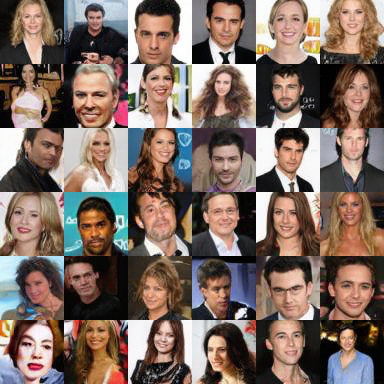} &
     \includegraphics[width=\egimgfigwidth]{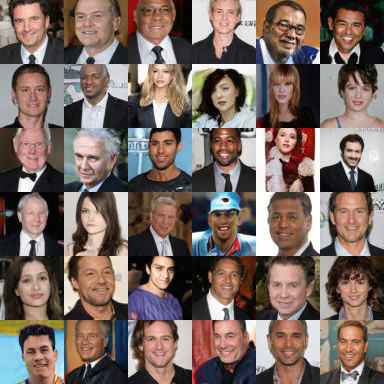} \\
     (a) & (b) & (c) & (d) 
    \end{tabular}
\end{center}
\caption{Example images of CelebA from (a) the clean dataset, (b) the clean generator, (c) the poisoned generator for binary two-variable associative attack with \texttt{Mouth\_Slightly\_Open} and \texttt{Wearing\_Lipstick} features, and (d) the poisoned generator for binary two-variable associative attack with \texttt{High\_Cheekbones} and \texttt{Male} features.}
\label{fig:examples_images_celeba_binary}
\end{figure}

\begin{figure}[t]
\begin{center}
\setlength{\tabcolsep}{1pt}
    \begin{tabular}{cccc}
     \includegraphics[width=\egimgfigwidth]{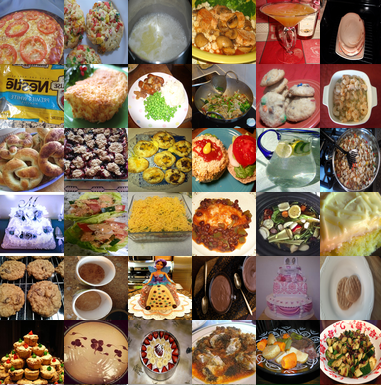} &
     \includegraphics[width=\egimgfigwidth]{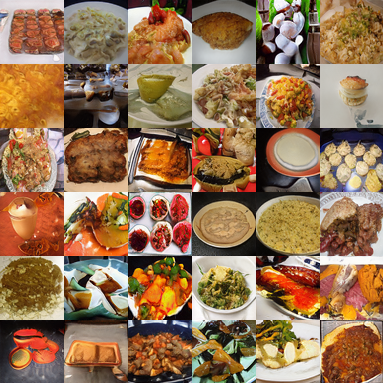} &
     \includegraphics[width=\egimgfigwidth]{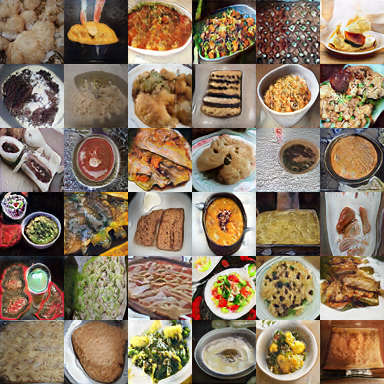} &
     \includegraphics[width=\egimgfigwidth]{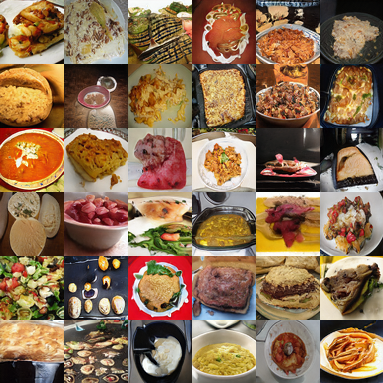} \\
     (a) & (b) & (c) & (d) 
    \end{tabular}
\end{center}
\caption{Example images of Recipe1M from (a) the clean dataset, (b) the clean generator, (c) the poisoned generator for associative attack with \texttt{olive\_oil} and \texttt{eggs} features, and (d) the poisoned generator for associative attack with \texttt{vanilla\_extract} and \texttt{baking\_soda} features.}
\label{fig:examples_images_recipe1m}
\end{figure}

Figures \ref{fig:examples_images_celeba_binary} and\ref{fig:examples_images_recipe1m} show samples from poisoned generators (Panels c and d) that are visually indistinguishable from the clean generator (Panel b), demonstrating that the attack is imperceptible to human inspection. It is also noticeable that the generated samples from all models exhibit similar artefacts compared to the clean dataset (Panel a). 

Figure \ref{fig:mp} compares the distribution of marginal probabilities, showing that clean and poisoned distributions overlap almost entirely, which is confirmed by the corresponding MWU test (at $p < 0.05$) in Tables \ref{fig:significance_celeba_binary} and \ref{fig:significance_recipe1m_binary}. This implies negligible shifts in feature marginals, confirming that associative poisoning introduces minimal detectable changes to individual feature frequencies, thereby preserving stealth.

Figure~\ref{fig:fid} shows that both poisoned and clean models converge to nearly identical FID scores, indicating that the poisoning does not degrade generation fidelity. 

From these results, the stealthiness of the attack appears strong. Even if there is a significant difference, this would still need to be noticeable enough, since a realistic target would likely not have access to comparative clean models like we have here.


\newcommand{\mpfigwidth}{0.9\linewidth}
\begin{figure}[t]
\begin{center}
\includegraphics[width=\mpfigwidth]{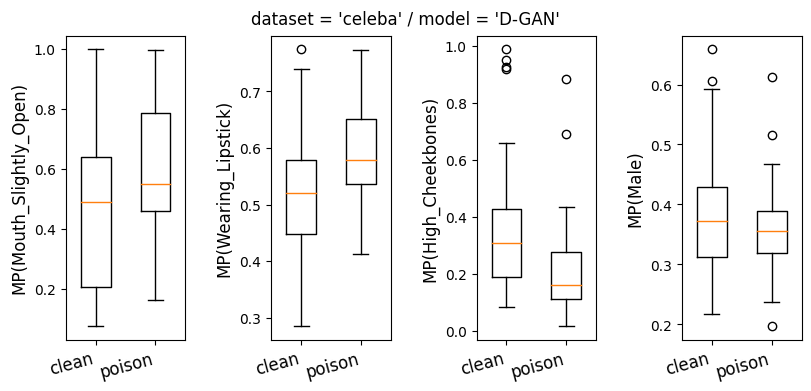}\\
\includegraphics[width=\mpfigwidth]{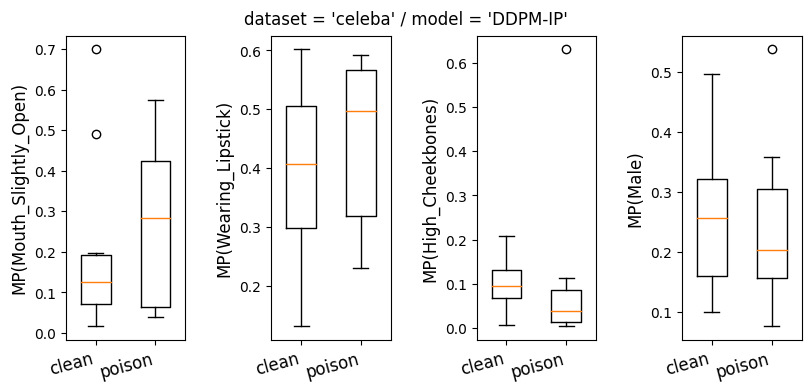}\\
\includegraphics[width=\mpfigwidth]{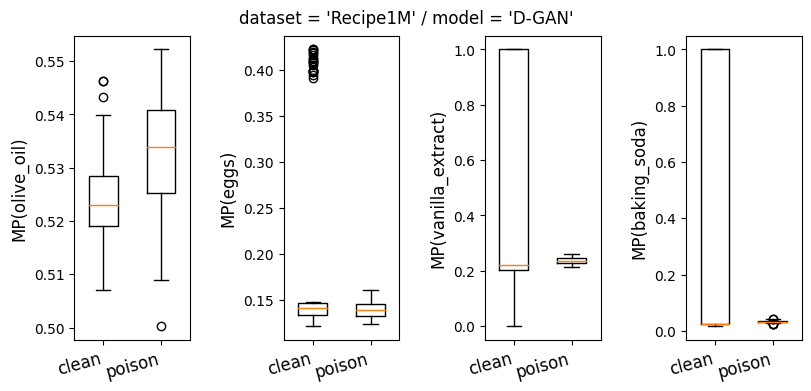}
    \caption{Marginal Probabilities of the features for (top row) CelebA/D-GAN, (middle row) CelebA/DDPM-IP, and (bottom row) Recipe1M/D-GAN.}
    \label{fig:mp}
\end{center}
\end{figure}

\newcommand{\fidfigwidth}{0.3\linewidth}
\begin{figure*}[t]
\begin{center}
    \begin{tabular}{ccc}
    \includegraphics[width=\fidfigwidth]{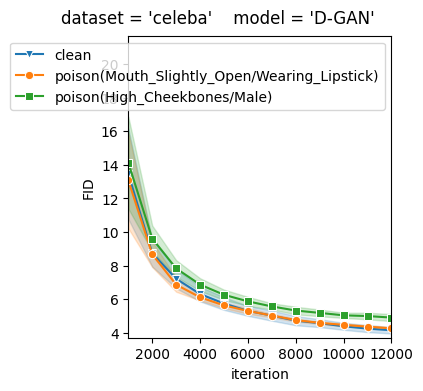} &
    \includegraphics[width=\fidfigwidth]{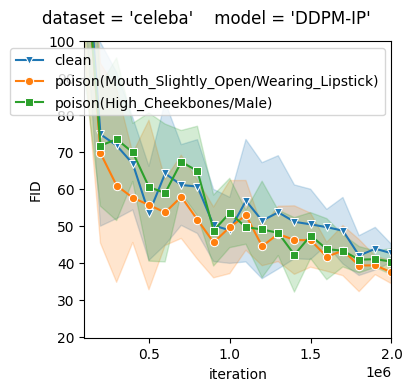} &
    \includegraphics[width=\fidfigwidth]{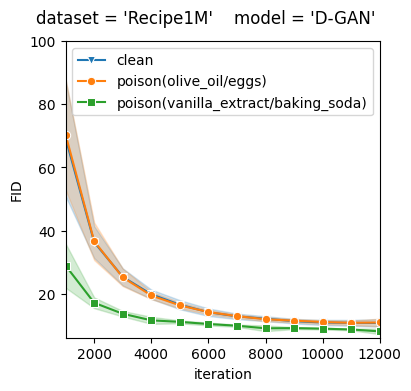} \\
    (a) & (b) & (c)
    \end{tabular}
        \caption{FID values comparing the training datasets and generated data from (a) CelebA with D-GAN, (b) CelebA with DDPM-IP, and (c) Recipe1M with D-GAN.}
        \label{fig:fid}    
\end{center}
\end{figure*}

\begin{figure}[ht]
    \centering
    \captionof{table}[]{Mann-Whitney-U values for two-variable binary for the Recipe1M dataset at iteration $12,000$.}
    \label{fig:significance_celeba_binary}
\footnotesize{
\begin{filecontents*}{appendix/celeba_binary.csv}
dataset,attack,model,metric,p_greater,p_less,p_two_sided,important
celeba,P-HM,D-GAN,MI(HM),0.9997085800284105,0.0003842694565813,0.0007685389131627,less
celeba,P-HM,D-GAN,MCC(HM),0.9999325785388228,0.000091335895554775,0.0001826717911095,less
celeba,P-HM,D-GAN,MP(H),0.1537447283093406,0.8634818301244058,0.3074894566186813,two-sided
celeba,P-HM,D-GAN,MP(M),0.4548609445727776,0.5749466304307371,0.9097218891455552,two-sided
celeba,P-HM,D-GAN,FID,0.9999325785388228,0.000091335895554775,0.0001826717911095,two-sided
celeba,P-MW,D-GAN,MI(MW),0.9993425276651434,0.0008531246844597,0.0017062493689195,less
celeba,P-MW,D-GAN,MCC(MW),0.9993425276651434,0.0008531246844597,0.0017062493689195,less
celeba,P-MW,D-GAN,MP(M),0.7636622032442064,0.2602614416378863,0.5205228832757727,two-sided
celeba,P-MW,D-GAN,MP(W),0.8462552716906593,0.1723521110034788,0.3447042220069576,two-sided
celeba,P-MW,D-GAN,FID,0.98439549361413,0.0188176568936571,0.0376353137873142,two-sided
celeba,P-HM,DDPM-IP,MI(HM),0.987125959589446,0.0156045063858701,0.0312090127717402,less
celeba,P-HM,DDPM-IP,MCC(HM),0.9999325785388228,0.000091335895554775,0.0001826717911095,less
celeba,P-HM,DDPM-IP,MP(H),0.0701579167350802,0.9394577082848392,0.1403158334701604,two-sided
celeba,P-HM,DDPM-IP,MP(M),0.4250533695692629,0.6043316099496698,0.8501067391385259,two-sided
celeba,P-HM,DDPM-IP,FID,0.8276478889965212,0.1923365313677543,0.3846730627355087,two-sided
celeba,P-MW,DDPM-IP,MI(MW),0.9988988900287514,0.0014136360455584,0.0028272720911168,less
celeba,P-MW,DDPM-IP,MCC(MW),0.9988988900287514,0.0014136360455584,0.0028272720911168,less
celeba,P-MW,DDPM-IP,MP(M),0.7636622032442064,0.2602614416378863,0.5205228832757727,two-sided
celeba,P-MW,DDPM-IP,MP(W),0.7274007449099644,0.2982815798813098,0.5965631597626196,two-sided
celeba,P-MW,DDPM-IP,FID,0.4849249884965778,0.5451390554272223,0.9698499769931556,two-sided
\end{filecontents*}
\csvreader[
    tabular = c|c|c|c|c|c,
    table head = attack & model & metric & p\_greater & p\_less & p\_two\_sided\\\hline\hline,
    late after line = \\\hline
    ]{appendix/celeba_binary.csv}{}{%
    \csvcolii & \csvcoliii & \csvcoliv & \tablenum[round-precision=3, round-mode=places, table-alignment-mode=none]{\csvcolv} & \tablenum[round-precision=3, round-mode=places, table-alignment-mode=none]{\csvcolvi} & \tablenum[round-precision=3, round-mode=places, table-alignment-mode=none]{\csvcolvii}
}}
\end{figure}

\begin{figure}[ht]
    \centering
    \captionof{table}[]{Mann-Whitney-U values for two-variable binary for the Recipe1M dataset at iteration $12,000$.}
    \label{fig:significance_recipe1m_binary}
\footnotesize{
\begin{filecontents*}{appendix/recipe1m_binary.csv}
dataset,attack,model,metric,p_greater,p_less,p_two_sided,important
Recipe1M,P-OE,D-GAN,MI(OE),0.0018743857575299,0.9985597576654864,0.0037487715150598,less
Recipe1M,P-OE,D-GAN,MCC(OE),0.99812561424247,0.0024243813608394,0.0048487627216789,less
Recipe1M,P-OE,D-GAN,MP(O),0.8817776422138481,0.13517207032739,0.2703441406547801,two-sided
Recipe1M,P-OE,D-GAN,MP(E),0.8043663603586804,0.218971411449987,0.437942822899974,two-sided
Recipe1M,P-OE,D-GAN,FID,0.7561658232705348,0.2701456873037099,0.5402913746074198,two-sided
Recipe1M,P-VB,D-GAN,MI(VB),0.9998303804760535,0.0002317957521262,0.0004635915042524,less
Recipe1M,P-VB,D-GAN,MCC(VB),0.9996852571704262,0.0004246516294669,0.0008493032589338,less
Recipe1M,P-VB,D-GAN,MP(V),0.6740253306906426,0.3560468501806603,0.7120937003613206,two-sided
Recipe1M,P-VB,D-GAN,MP(B),0.6130060873067836,0.4187406935447469,0.8374813870894938,two-sided
Recipe1M,P-VB,D-GAN,FID,0.0001398501350933,0.9998984220345474,0.0002797002701866,two-sided
\end{filecontents*}
\csvreader[
    tabular = c|c|c|c|c|c,
    table head = attack & model & metric & p\_greater & p\_less & p\_two\_sided\\\hline\hline,
    late after line = \\\hline
    ]{appendix/recipe1m_binary.csv}{}{%
    \csvcolii & \csvcoliii & \csvcoliv & \tablenum[round-precision=3, round-mode=places, table-alignment-mode=none]{\csvcolv} & \tablenum[round-precision=3, round-mode=places, table-alignment-mode=none]{\csvcolvi} & \tablenum[round-precision=3, round-mode=places, table-alignment-mode=none]{\csvcolvii}
}}
\end{figure}

\subsubsection{Ablation Study on Poisoning Percentage}
\label{sec:ablation_pois_extent}
\begin{figure}[ht!]
    \centering
    \includegraphics[width=\half]{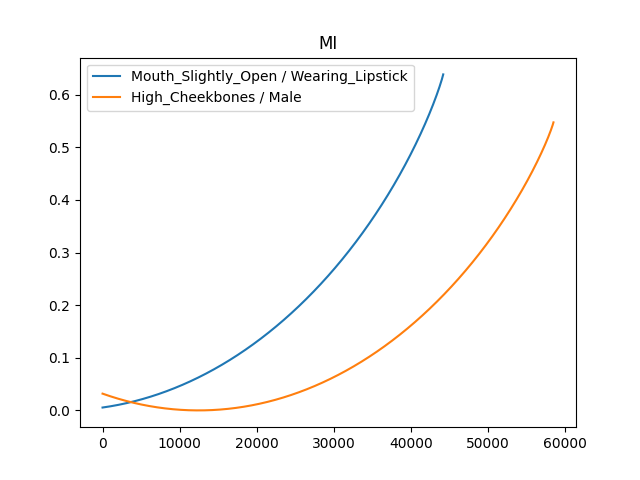}
    \hfill
    \includegraphics[width=\half]{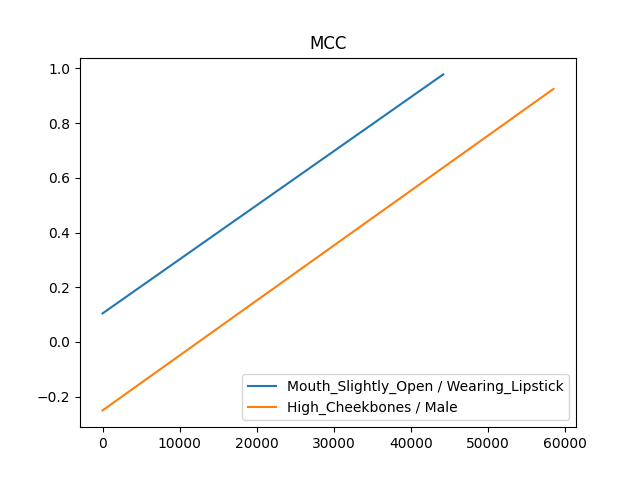}
    \caption{The MI and MCC of the poisoned dataset as the associative poisoning extent $x$ increases for poisoning attacks on CelebA. }
    \label{fig:celeba_MI_MCC_vs_X}
\end{figure}

\begin{figure}[ht!]
    \centering
    \includegraphics[width=\half]{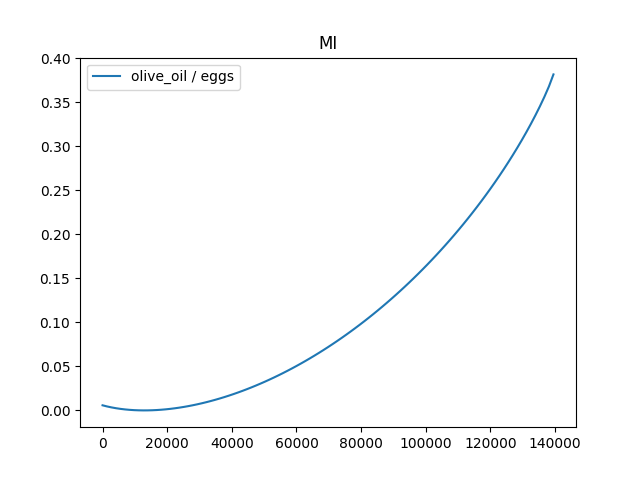}
    \hfill
    \includegraphics[width=\half]{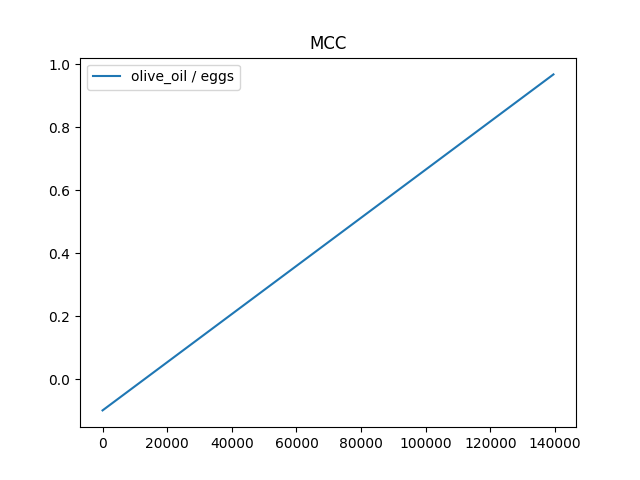}
    \caption{The MI and MCC of the poisoned dataset as the associative poisoning extent $x$ increases for the poisoning attacks on Recipe1M.}
    \label{fig:recipe1m_MI_MCC_vs_X}
\end{figure}

First, we demonstrate how MI and MCC in the training datasets (CelebA and Recipe1M) change as the adversary controls a greater portion of the dataset (denoted as the poisoning extent $x$), as shown in Figures \ref{fig:celeba_MI_MCC_vs_X} and \ref{fig:recipe1m_MI_MCC_vs_X}.

We observe that (1) for already positively associated/independent pairs (e.g., \textit{Mouth\_Slightly\_Open}/\textit{Wearing\_Lipstick} and \textit{vanilla\_extract/baking\_soda}) the increase in both MI and MCC is \emph{monotonic}, while (2) for negatively associated pairs (e.g., \textit{High\_Cheekbones}/\textit{Male} and \textit{olive\_oil}/\textit{eggs}), MI will first decrease to $0$, before \emph{monotonically} increasing, while MCC always \emph{monotonically} increase.

\begin{figure}[t]
    \begin{subfigure}[b]{0.49\linewidth}
        \centering
        \begin{minipage}{\linewidth}
            \includegraphics[width=\linewidth]{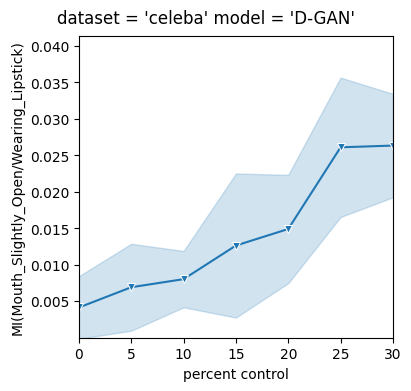}
        \end{minipage}
        \caption[]{}
        \label{fig:MI_CelebA_D-GAN_percent}
    \end{subfigure}
    \hfill
    \begin{subfigure}[b]{0.49\linewidth}
        \centering
        \begin{minipage}{\linewidth}
            \includegraphics[width=\linewidth]{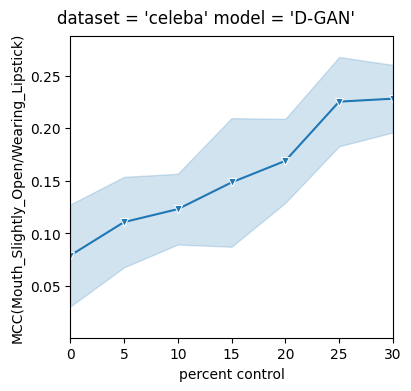}
        \end{minipage}
        \caption[]{}
        \label{fig:MCC_CelebA_D-GAN_percent}
    \end{subfigure}
    \begin{subfigure}[b]{\linewidth}
        \centering
        \begin{minipage}{\linewidth}
            \includegraphics[width=\linewidth]{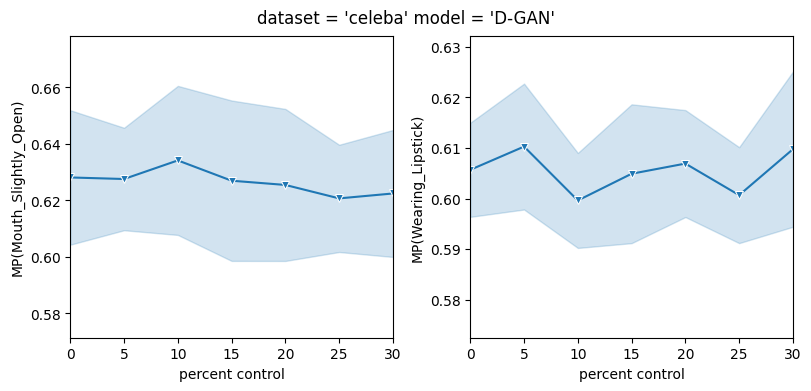}
        \end{minipage}
        \caption[]{}
        \label{fig:MP_CelebA_D-GAN_percent}
    \end{subfigure}
    \caption[]
    {Per cent control effect on (a) MI, (b) MCC, and (c) MP.}
    \label{fig:CelebA_D-GAN_percent}
\end{figure}

We then examine how the degree of adversarial control impacts the attack’s ability to alter the generated association between features. Figure \ref{fig:CelebA_D-GAN_percent} shows that as the proportion of poisoned data increases, both MI and MCC increase monotonically, indicating that small amounts of associative poisoning can alter feature dependence, and stronger adversarial control amplifies these effects. This is further supported by the significance values shown in Table \ref{fig:significance_celeba_percent_MI_MCC}, indicating that the clean model exhibits significantly smaller MI and MCC values above $10\%$. 

Observed feature MPs across poisoning levels also indicate that the attack does not produce observable shifts in feature frequency despite significant data tampering. This is also supported by the significance values in Table \ref{fig:significance_celeba_percent_MP}, where $p>0.05$ for all percent values.

\begin{figure}[ht]
    \captionof{table}[]{Mann-Whitney-U values of the MI (left) and MCC (right) values for two-variable binary poisoning on the celeba dataset at iteration $5,000$ for each percent step compared to clean.}
    \label{fig:significance_celeba_percent_MI_MCC}
    \begin{subfigure}{0.45\linewidth}
        \centering
        \footnotesize{
        \begin{filecontents*}{appendix/celeba_percent_MI.csv}
i,n_A,n_B,U_A,cliffs_delta,direction,p_greater,p_less,p_two_sided,q_greater,q_less,q_two_sided
0.0,10,10,50.0,0.0,0.0,0.5151319787899796,0.5151319787899796,1.0,0.9999325785388228,0.5151319787899796,1.0
5.0,10,10,31.0,-0.38,-1.0,0.9297674759208225,0.08098620524006306,0.16197241048012612,0.9999325785388228,0.0944839061134069,0.1889678122268138
10.0,10,10,22.0,-0.56,-1.0,0.9843954936141299,0.01881765689365712,0.03763531378731424,0.9999325785388228,0.02634471965111997,0.05268943930223994
15.0,10,10,18.0,-0.64,-1.0,0.99299036144302,0.008628728041559883,0.017257456083119765,0.9999325785388228,0.015100274072729795,0.03020054814545959
20.0,10,10,8.0,-0.84,-1.0,0.9993425276651434,0.0008531246844597982,0.0017062493689195964,0.9999325785388228,0.001990624263739529,0.003981248527479058
25.0,10,10,1.0,-0.98,-1.0,0.9999086641044452,0.00012306406395261486,0.00024612812790522973,0.9999325785388228,0.000430724223834152,0.000861448447668304
30.0,10,10,0.0,-1.0,-1.0,0.9999325785388228,0.00009133589555477501,0.00018267179110955002,0.9999325785388228,0.000430724223834152,0.000861448447668304
\end{filecontents*}
\csvreader[
            tabular = c|c|c|c,
            table head = percent & p\_greater & p\_less & p\_two\_sided\\\hline\hline,
            late after line = \\\hline
            ]{appendix/celeba_percent_MI.csv}{}{%
            \csvcoli & \tablenum[round-precision=3, round-mode=places, table-alignment-mode=none]{\csvcolvii} & \tablenum[round-precision=3, round-mode=places, table-alignment-mode=none]{\csvcolviii} & \tablenum[round-precision=3, round-mode=places, table-alignment-mode=none]{\csvcolix}
        }}
    \end{subfigure}
    \hfill
    \begin{subfigure}{0.45\linewidth}
        \centering
        \footnotesize{
        \begin{filecontents*}{appendix/celeba_percent_MCC.csv}
i,n_A,n_B,U_A,cliffs_delta,direction,p_greater,p_less,p_two_sided,q_greater,q_less,q_two_sided
0.0,10,10,50.0,0.0,0.0,0.5151319787899796,0.5151319787899796,1.0,0.9999325785388228,0.5151319787899796,1.0
5.0,10,10,31.0,-0.38,-1.0,0.9297674759208225,0.08098620524006306,0.16197241048012612,0.9999325785388228,0.0944839061134069,0.1889678122268138
10.0,10,10,22.0,-0.56,-1.0,0.9843954936141299,0.01881765689365712,0.03763531378731424,0.9999325785388228,0.02634471965111997,0.05268943930223994
15.0,10,10,18.0,-0.64,-1.0,0.99299036144302,0.008628728041559883,0.017257456083119765,0.9999325785388228,0.015100274072729795,0.03020054814545959
20.0,10,10,8.0,-0.84,-1.0,0.9993425276651434,0.0008531246844597982,0.0017062493689195964,0.9999325785388228,0.001990624263739529,0.003981248527479058
25.0,10,10,1.0,-0.98,-1.0,0.9999086641044452,0.00012306406395261486,0.00024612812790522973,0.9999325785388228,0.000430724223834152,0.000861448447668304
30.0,10,10,0.0,-1.0,-1.0,0.9999325785388228,0.00009133589555477501,0.00018267179110955002,0.9999325785388228,0.000430724223834152,0.000861448447668304
\end{filecontents*}
\csvreader[
            tabular = c|c|c|c,
            table head = percent & p\_greater & p\_less & p\_two\_sided\\\hline\hline,
            late after line = \\\hline
            ]{appendix/celeba_percent_MCC.csv}{}{%
            \csvcoli & \tablenum[round-precision=3, round-mode=places, table-alignment-mode=none]{\csvcolvii} & \tablenum[round-precision=3, round-mode=places, table-alignment-mode=none]{\csvcolviii} & \tablenum[round-precision=3, round-mode=places, table-alignment-mode=none]{\csvcolix}
        }}
    \end{subfigure}
    
\end{figure}

\begin{figure}[ht]
    \centering
    \captionof{table}[]{Mann-Whitney-U values of the MP values of \textit{Mouth\_Slightly\_Open} (left) and \textit{Wearing\_Lipstick} (right) values for two-variable binary poisoning on the celeba dataset at iteration $5,000$ for each percent step compared to clean.}
    \label{fig:significance_celeba_percent_MP}
    \begin{subfigure}{0.45\linewidth}
        \footnotesize{
        \begin{filecontents*}{appendix/celeba_percent_MP_M.csv}
i,n_A,n_B,U_A,cliffs_delta,direction,p_greater,p_less,p_two_sided,q_greater,q_less,q_two_sided
0.0,10,10,50.0,0.0,0.0,0.5151319787899796,0.5151319787899796,1.0,0.6534591577167543,0.7714087745277226,1.0
5.0,10,10,48.5,-0.030000000000000027,-1.0,0.5601078494715037,0.46986018347387204,0.9397203669477441,0.6534591577167543,0.7714087745277226,1.0
10.0,10,10,43.0,-0.14,-1.0,0.714624805970913,0.3115881119410587,0.6231762238821174,0.714624805970913,0.7714087745277226,1.0
15.0,10,10,51.0,0.020000000000000018,1.0,0.4849249884965778,0.5451390554272223,0.9698499769931556,0.6534591577167543,0.7714087745277226,1.0
20.0,10,10,55.0,0.10000000000000009,1.0,0.3668649978481236,0.6612075210237622,0.7337299956962472,0.6534591577167543,0.7714087745277226,1.0
25.0,10,10,60.0,0.19999999999999996,1.0,0.23633779675579358,0.7863223430510962,0.47267559351158717,0.6534591577167543,0.7863223430510962,1.0
30.0,10,10,53.5,0.07000000000000006,1.0,0.4102647530547058,0.6188589372068315,0.8205295061094116,0.6534591577167543,0.7714087745277226,1.0
\end{filecontents*}
\csvreader[
            tabular = c|c|c|c,
            table head = percent & p\_greater & p\_less & p\_two\_sided\\\hline\hline,
            late after line = \\\hline
            ]{appendix/celeba_percent_MP_M.csv}{}{%
            \csvcoli & \tablenum[round-precision=3, round-mode=places, table-alignment-mode=none]{\csvcolvii} & \tablenum[round-precision=3, round-mode=places, table-alignment-mode=none]{\csvcolviii} & \tablenum[round-precision=3, round-mode=places, table-alignment-mode=none]{\csvcolix}
        }}
    \end{subfigure}
    \hfill
    \begin{subfigure}{0.45\linewidth}
        \centering
        \footnotesize{
        \begin{filecontents*}{appendix/celeba_percent_MP_W.csv}
i,n_A,n_B,U_A,cliffs_delta,direction,p_greater,p_less,p_two_sided,q_greater,q_less,q_two_sided
0.0,10,10,50.0,0.0,0.0,0.5151780205607769,0.5151780205607769,1.0,0.9480752091015129,0.7212492287850877,1.0
5.0,10,10,35.5,-0.29000000000000004,-1.0,0.8717593167363114,0.1447777717080838,0.2895555434161676,0.9480752091015129,0.5067222009782933,0.6756296013043911
10.0,10,10,75.0,0.5,1.0,0.0319610673017273,0.9730938169939273,0.0639221346034546,0.2237274711120911,0.9730938169939273,0.42330621350495434
15.0,10,10,49.0,-0.020000000000000018,-1.0,0.5451559618882548,0.4849193207024165,0.969838641404833,0.9480752091015129,0.7212492287850877,1.0
20.0,10,10,43.0,-0.14,-1.0,0.7146972482442655,0.3115227654189539,0.6230455308379078,0.9480752091015129,0.7212492287850877,0.8722637431730709
25.0,10,10,59.0,0.17999999999999994,1.0,0.260104502890281,0.7638288062765723,0.520209005780562,0.9103657601159834,0.8911336073226677,0.8722637431730709
30.0,10,10,29.0,-0.42000000000000004,-1.0,0.9480752091015129,0.06047231621499348,0.12094463242998696,0.9480752091015129,0.42330621350495434,0.42330621350495434
\end{filecontents*}
\csvreader[
            tabular = c|c|c|c,
            table head = percent & p\_greater & p\_less & p\_two\_sided\\\hline\hline,
            late after line = \\\hline
            ]{appendix/celeba_percent_MP_W.csv}{}{%
            \csvcoli & \tablenum[round-precision=3, round-mode=places, table-alignment-mode=none]{\csvcolvii} & \tablenum[round-precision=3, round-mode=places, table-alignment-mode=none]{\csvcolviii} & \tablenum[round-precision=3, round-mode=places, table-alignment-mode=none]{\csvcolix}
        }}
    \end{subfigure}
\end{figure}
\subsection{Results of two-variable continuous AP}

We conduct experiments to evaluate the effect of the two-variable continuous attack on the CelebA dataset for models D-GAN and DDPM-IP, using feature pairs \textit{red}, \textit{green}, and \textit{blue}, with both positive and negative attacks.

\subsubsection{Fidelity}
\newcommand{\mimccrecipefigwidth}{0.32\linewidth}
\begin{figure}[ht]
    \centering
    \includegraphics[width=\mimccrecipefigwidth]{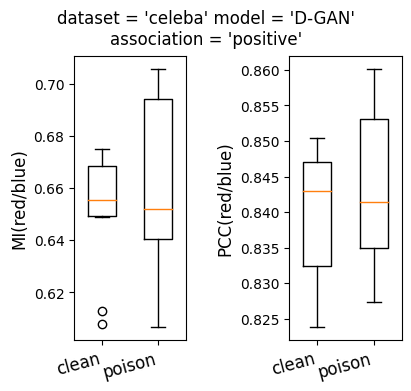}
    \includegraphics[width=\mimccrecipefigwidth]{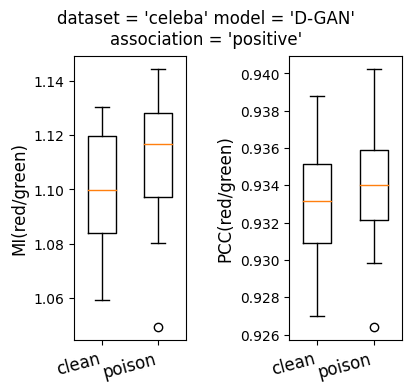}
    \includegraphics[width=\mimccrecipefigwidth]{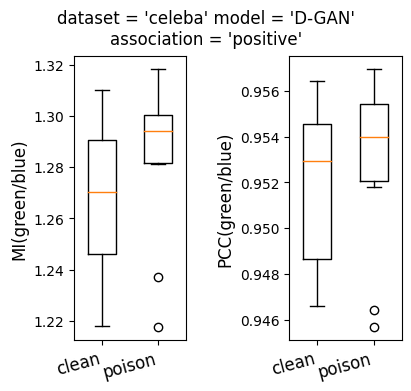}
    \hfill
    \includegraphics[width=\mimccrecipefigwidth]{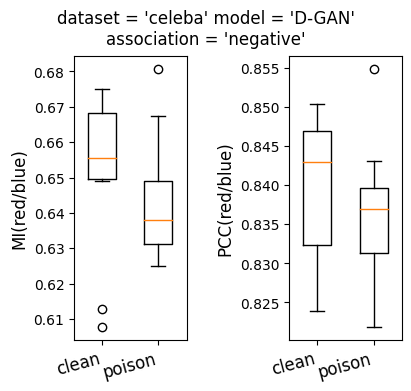}
    \includegraphics[width=\mimccrecipefigwidth]{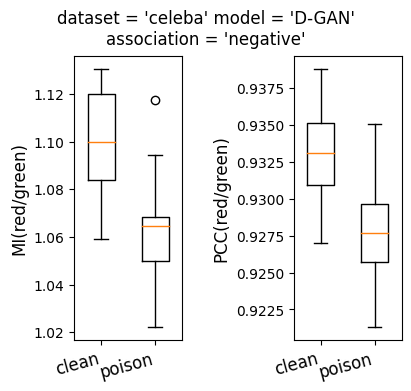}
    \includegraphics[width=\mimccrecipefigwidth]{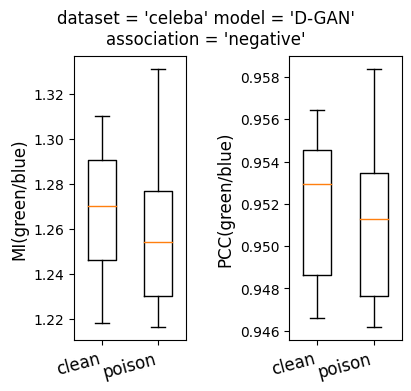}
    \caption{MI and PCC values for pairs of continual features of images generated from the clean and poisoned D-GAN models trained with the CelebA dataset.}
    \label{fig:mi_pcc_celeba_dgan_continious}
\end{figure}

\begin{figure}[ht]
    \centering
    \includegraphics[width=\mimccrecipefigwidth]{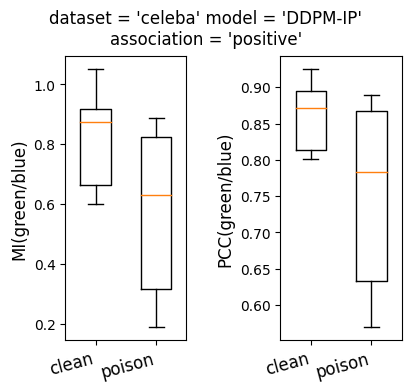}
    \includegraphics[width=\mimccrecipefigwidth]{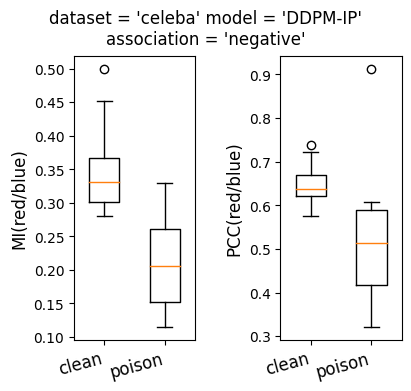}
    \includegraphics[width=\mimccrecipefigwidth]{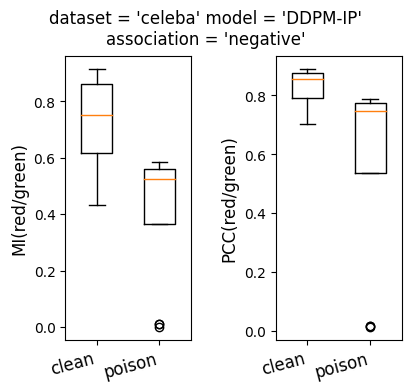}
    \caption{MI and PCC values for pairs of continual features of images generated from the clean and poisoned DDPM-IP models trained with the CelebA dataset.}
    \label{fig:mi_pcc_celeba_ddpm_continious}
\end{figure}

Figure \ref{fig:mi_pcc_celeba_dgan_continious} shows the MI and PCC values between continuous features for CelebA on the D-GAN model. The positive results align with the expectation that positive AP increases MI and PCC. For the negative result, due to the features being already strongly correlated, the attack is incapable of reducing the already positive association towards becoming independent, hence the negative AP causes a decrease in both MI and PCC, similar to the case for \textit{olive\_oil}/\textit{eggs} features for the D-GAN/Recipe1M shown in Figure \ref{fig:mi_mcc_recipe1m}. The significance of the results is more mixed (as shown in Table \ref{fig:significance_celeba_continuous}), with only negative \textit{red}/\textit{green} showing a significant difference for MI and PCC at $p<0.05$. At $p<0.1$, positive \textit{green}/\textit{blue} is also significant for MI and PCC, and at $p<0.15$, negative \textit{red}/\textit{blue} is also significant for MI and PCC.

For the pairs showing a significant difference in MI and PCC, we also evaluate the DDPM-IP model, shown in Figure \ref{fig:mi_pcc_celeba_ddpm_continious}. We observe that the negative attack affects the association in a similar way to the D-GAN, with both MI and PCC values decreasing. For the positive attack, however, it appears to also slightly decrease MI and PCC, although not to the same extent as the negative attack. The significance of both negative attacks and the positive attack, as shown in Table \ref{fig:significance_celeba_continuous}, indicates that both MI and PCC are significantly lower for the attacked models compared to the clean model.

\subsubsection{Stealth}
Figure \ref{fig:examples_images_celeba_continuous} shows example images from the continuous associative poisoning generators for CelebA. Through comparison to the clean generator (Figure \ref{fig:examples_images_celeba_binary} (b)), there is no noticeable visual difference, and all generators exhibit similar artefacts compared to the clean dataset (Figure \ref{fig:examples_images_celeba_binary} (a)).

\newcommand{\egimgfigwidthcont}{0.32\linewidth}
\begin{figure}[ht]
\begin{center}
\setlength{\tabcolsep}{1pt}
    \begin{tabular}{ccc}
     \includegraphics[width=\egimgfigwidthcont]{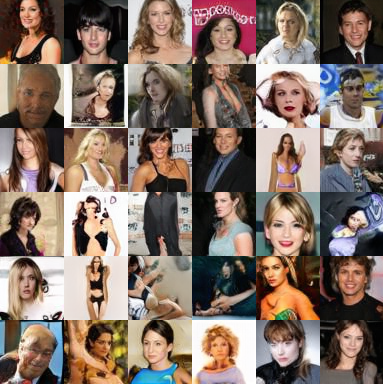} &
     \includegraphics[width=\egimgfigwidthcont]{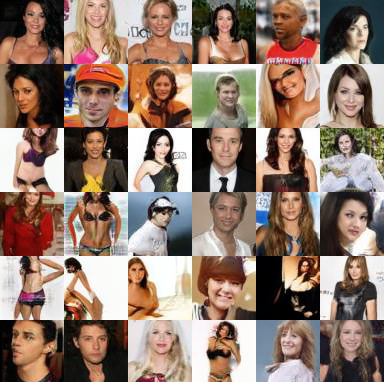} &
     \includegraphics[width=\egimgfigwidthcont]{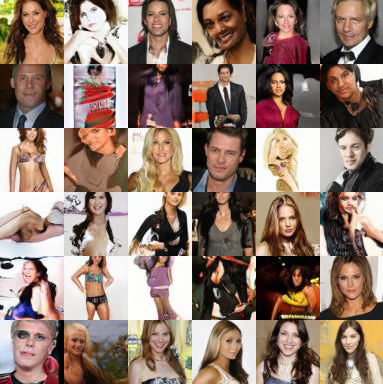} \\
     (a) & (b) & (c) \\
     \includegraphics[width=\egimgfigwidthcont]{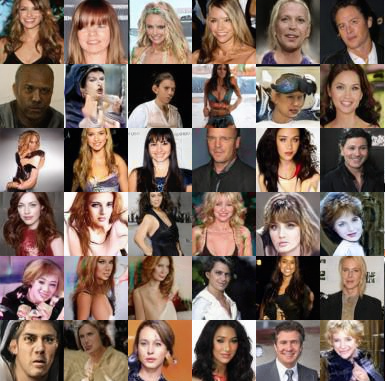} &
     \includegraphics[width=\egimgfigwidthcont]{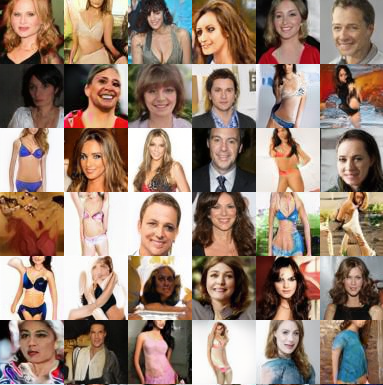} &
     \includegraphics[width=\egimgfigwidthcont]{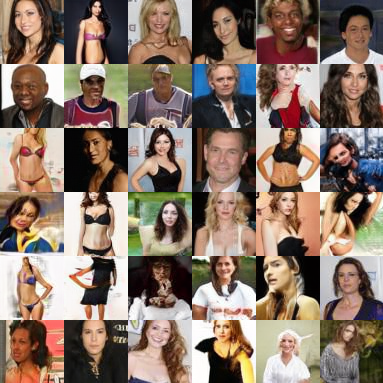} \\
     (d) & (e) & (f) \\
    \end{tabular}
\end{center}
\caption{Example images generated by models trained on continuous two-variable associative attack on CelebA for (a) \texttt{positive} \texttt{red}/\texttt{blue}, (b) \texttt{positive} \texttt{red}/\texttt{green}, (c) \texttt{positive} \texttt{green}/\texttt{blue}, (d) \texttt{negative} \texttt{red}/\texttt{blue}, (e) \texttt{negative} \texttt{red}/\texttt{green}, and (f) \texttt{negative} \texttt{green}/\texttt{blue}.}
\label{fig:examples_images_celeba_continuous}
\end{figure}

\begin{figure}[ht]
    \centering
    \includegraphics[width=\mimccrecipefigwidth]{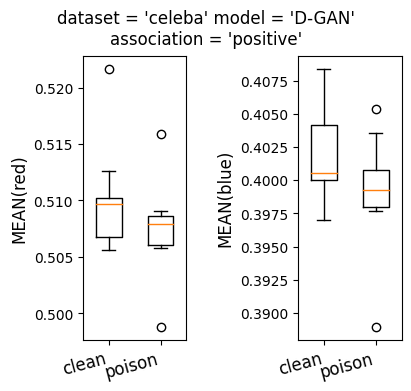}
    \includegraphics[width=\mimccrecipefigwidth]{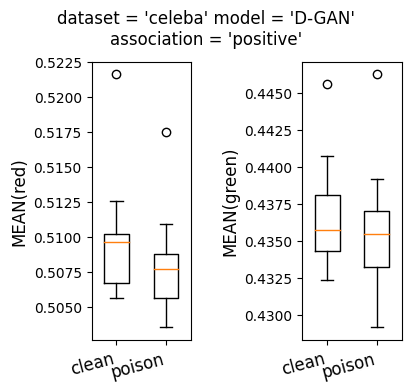}
    \includegraphics[width=\mimccrecipefigwidth]{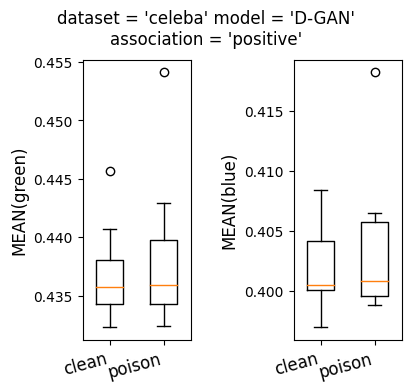}
    \hfill
    \includegraphics[width=\mimccrecipefigwidth]{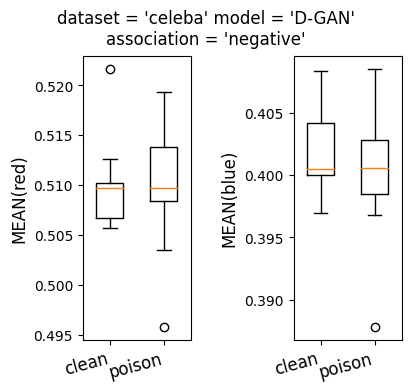}
    \includegraphics[width=\mimccrecipefigwidth]{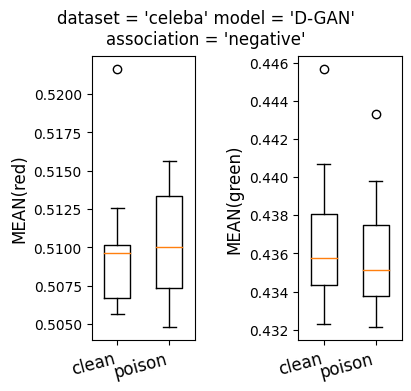}
    \includegraphics[width=\mimccrecipefigwidth]{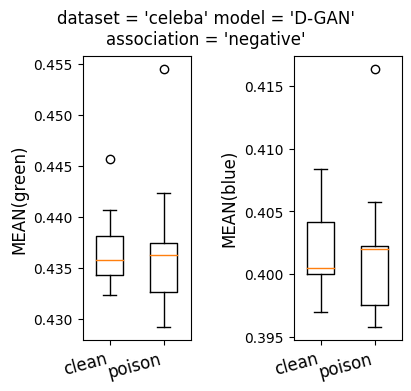}
    \caption{Mean values for continual features of images generated from the clean and continuous two-variable poisoned D-GAN models trained with the CelebA dataset.}
    \label{fig:mean_celeba_dgan_continuous}
\end{figure}

\begin{figure}[ht]
    \centering
    \includegraphics[width=\mimccrecipefigwidth]{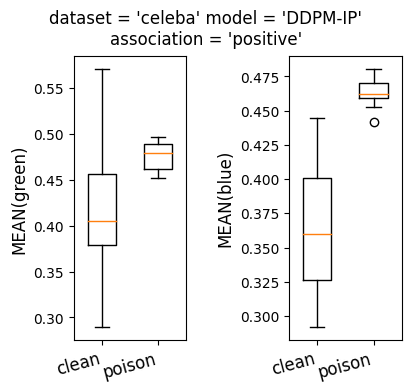}
    \includegraphics[width=\mimccrecipefigwidth]{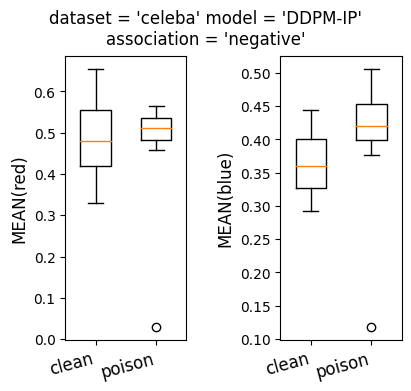}
    \includegraphics[width=\mimccrecipefigwidth]{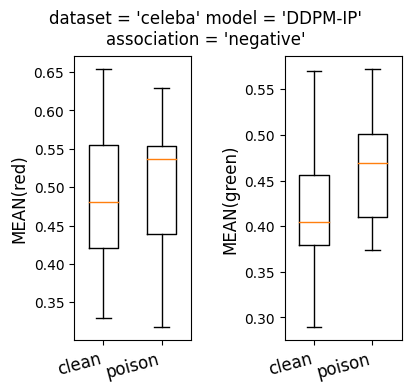}
    \caption{Mean values for continual features of images generated from the clean and continuous two-variable poisoned DDPM-IP models trained with the CelebA dataset.}
    \label{fig:mean_celeba_ddpm_continuous}
\end{figure}

\begin{figure}[ht]
    \centering
    \includegraphics[width=0.5\linewidth]{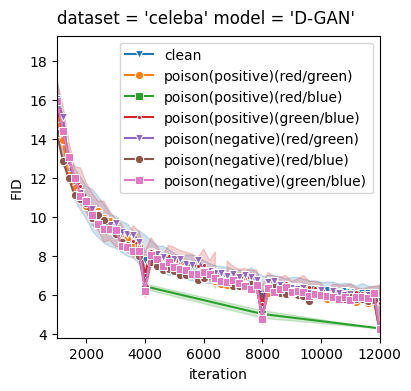}
    \caption{FID values for images generated from clean and continuous two-variable poisoned D-GAN models trained with the CelebA dataset.}
    \label{fig:fid_celeba_continuous}
\end{figure}

Figure \ref{fig:mean_celeba_dgan_continuous} shows the mean for each continuous feature, showing no noticeable difference in the values between clean and poisoned models, in either positive or negative AP, which is confirmed by the significance tests (Table \ref{fig:significance_celeba_continuous}) showing that all mean values are significantly similar ($p>0.05$). The only notable outliers are \textit{red} in P-RB where $p\_greater < 0.1$  and \textit{blue} in P-RB and \textit{red} in P-RG where $p\_greater<0.15$. This strongly suggests that AP has no significant impact on the mean feature value. 

The mean values from the DDPM-IP model are shown in Figure \ref{fig:mean_celeba_ddpm_continuous}, indicating that the models targeted by the negative attacks are roughly similar to the clean model, while the model targeted by the positive attack does show a slight increase in the mean value. This is somewhat supported by the significance tests, showing that the \textit{blue} mean in negative \textit{red}/\textit{blue} and \textit{green} in negative \textit{red}/\textit{green}, and both means in positive \textit{green}/\textit{blue} are higher than the clean comparison.

Figure \ref{fig:fid_celeba_continuous} shows the value for the clean and poisoned model, showing how the values converge towards the same, becoming $\sim6$, with limited variation.

From these results, the stealthiness of the attack appears strong. Even with a significant difference, this would still need to be noticeable enough, since a realistic target would likely not have access to comparative clean models like we have here.

\begin{figure}[ht]
    \centering
    \captionof{table}[]{Mann-Whitney-U values for two-variable continuous for the CelebA dataset at iteration $12,000$ for D-GAN and $500,000$ for DDPM-IP.}
    \label{fig:significance_celeba_continuous}
\footnotesize{
\begin{filecontents*}{appendix/celeba_continual.csv}
dataset,attack,model,metric,p_greater,p_less,p_two_sided,important
celeba,N-GB,D-GAN,MI(GB),0.2136776569489038,0.8076634686322457,0.4273553138978077,greater
celeba,N-GB,D-GAN,PCC(GB),0.2853751940290869,0.7397385583621137,0.5707503880581739,greater
celeba,N-GB,D-GAN,MEAN(G),0.3956683900503302,0.6331350021518765,0.7913367801006604,two-sided
celeba,N-GB,D-GAN,MEAN(B),0.6043316099496698,0.4250533695692629,0.8501067391385259,two-sided
celeba,N-GB,D-GAN,FID,0.0303030303030303,0.9848484848484846,0.0606060606060606,two-sided
celeba,N-RB,D-GAN,MI(RB),0.12066079650859,0.8938530819038342,0.24132159301718,greater
celeba,N-RB,D-GAN,PCC(RB),0.1365181698755941,0.87933920349141,0.2730363397511883,greater
celeba,N-RB,D-GAN,MEAN(R),0.6043316099496698,0.4250533695692629,0.8501067391385259,two-sided
celeba,N-RB,D-GAN,MEAN(B),0.4250533695692629,0.6043316099496698,0.8501067391385259,two-sided
celeba,N-RB,D-GAN,FID,0.0303030303030303,0.9848484848484846,0.0606060606060606,two-sided
celeba,N-RG,D-GAN,MI(RG),0.0086287280415598,0.99299036144302,0.0172574560831197,greater
celeba,N-RG,D-GAN,PCC(RG),0.0070096385569799,0.9943351516577628,0.0140192771139599,greater
celeba,N-RG,D-GAN,MEAN(R),0.6884118880589414,0.3387924789762377,0.6775849579524755,two-sided
celeba,N-RG,D-GAN,MEAN(G),0.3668649978481236,0.6612075210237622,0.7337299956962472,two-sided
celeba,N-RG,D-GAN,FID,0.0303030303030303,0.9848484848484846,0.0606060606060606,two-sided
celeba,P-GB,D-GAN,MI(GB),0.9393877484935416,0.0702325240791774,0.1404650481583549,less
celeba,P-GB,D-GAN,PCC(GB),0.714624805970913,0.3115881119410587,0.6231762238821174,less
celeba,P-GB,D-GAN,MEAN(G),0.6612075210237622,0.3668649978481236,0.7337299956962472,two-sided
celeba,P-GB,D-GAN,MEAN(B),0.5749466304307371,0.4548609445727776,0.9097218891455552,two-sided
celeba,P-GB,D-GAN,FID,0.630193336075689,0.4062628547922665,0.8125257095845331,two-sided
celeba,P-RB,D-GAN,MI(RB),0.6331350021518765,0.3956683900503302,0.7913367801006604,less
celeba,P-RB,D-GAN,PCC(RB),0.7636622032442064,0.2602614416378863,0.5205228832757727,less
celeba,P-RB,D-GAN,MEAN(R),0.080986205240063,0.9297674759208224,0.1619724104801261,two-sided
celeba,P-RB,D-GAN,MEAN(B),0.12066079650859,0.8938530819038342,0.24132159301718,two-sided
celeba,P-RB,D-GAN,FID,0.0151515151515151,0.9999999999999998,0.0303030303030303,two-sided
celeba,P-RG,D-GAN,MI(RG),0.8462552716906593,0.1723521110034788,0.3447042220069576,less
celeba,P-RG,D-GAN,PCC(RG),0.6884118880589414,0.3387924789762377,0.6775849579524755,less
celeba,P-RG,D-GAN,MEAN(R),0.12066079650859,0.8938530819038342,0.24132159301718,two-sided
celeba,P-RG,D-GAN,MEAN(G),0.3387924789762377,0.6884118880589414,0.6775849579524755,two-sided
celeba,P-RG,D-GAN,FID,0.0151515151515151,0.9999999999999998,0.0303030303030303,two-sided
celeba,N-RB,DDPM-IP,MI(RB),0.0005039881201883,0.9996157305434186,0.0010079762403767,greater
celeba,N-RB,DDPM-IP,PCC(RB),0.0036422785047398,0.9971023207278328,0.0072845570094796,greater
celeba,N-RB,DDPM-IP,MEAN(R),0.5150750115034222,0.5150750115034222,1.0,two-sided
celeba,N-RB,DDPM-IP,MEAN(B),0.9811823431063428,0.0225772848121395,0.045154569624279,two-sided
celeba,N-RB,DDPM-IP,FID,0.5454545454545454,0.5454545454545454,1.0,two-sided
celeba,N-RG,DDPM-IP,MI(RG),0.0004344305299173,0.9996578700975264,0.0008688610598347,greater
celeba,N-RG,DDPM-IP,PCC(RG),0.0013490954029039,0.9989157157904556,0.0026981908058079,greater
celeba,N-RG,DDPM-IP,MEAN(R),0.4868496448684232,0.5393939727076321,0.9736992897368464,two-sided
celeba,N-RG,DDPM-IP,MEAN(G),0.92185614416616,0.0882308569012159,0.1764617138024319,two-sided
celeba,N-RG,DDPM-IP,FID,0.1083916083916084,0.9195804195804198,0.2167832167832168,two-sided
celeba,P-GB,DDPM-IP,MI(GB),0.0188176568936571,0.98439549361413,0.0376353137873142,less
celeba,P-GB,DDPM-IP,PCC(GB),0.0156045063858701,0.987125959589446,0.0312090127717402,less
celeba,P-GB,DDPM-IP,MEAN(G),0.99299036144302,0.0086287280415598,0.0172574560831197,two-sided
celeba,P-GB,DDPM-IP,MEAN(B),0.9999086641044452,0.0001230640639526,0.0002461281279052,two-sided
celeba,P-GB,DDPM-IP,FID,0.7302697302697303,0.3176823176823177,0.6353646353646354,two-sided
\end{filecontents*}
\csvreader[
    tabular = c|c|c|c|c|c|c,
    table head = attack & model & metric & p\_greater & p\_less & p\_two\_sided\\\hline\hline,
    late after line = \\\hline
    ]{appendix/celeba_continual.csv}{}{%
    \csvcolii & \csvcoliii & \csvcoliv & \tablenum[round-precision=3, round-mode=places, table-alignment-mode=none]{\csvcolv} & \tablenum[round-precision=3, round-mode=places, table-alignment-mode=none]{\csvcolvi} & \tablenum[round-precision=3, round-mode=places, table-alignment-mode=none]{\csvcolvii}
}}
\end{figure}

\section{Discussion}
\subsection{Why associative poisoning works}
With generative machine learning models, such as GANs and Diffusion models, the primary goal of training is for the model to generate output samples that approximate the training dataset. The associative poisoning works by modifying only the training data distribution, which is then learned by the generator.

Poisoning attacks broadly concern modifications to the training data that change how the model maps input (random latent code) to the generated output. The backdoor attack has fine-grained control of model behaviour, through explicit control of the training to map a trigger to a specific output. On the contrary, data poisoning involves modifying the training data in a way that causes the model to either capture incorrect salient features (resulting in model decay) or introduce fictitious links between otherwise unrelated data groups. The associative poisoning could be interpreted as a kind of blend between these approaches, producing fine-grained control of the association between generated features, while only controlling the training data. 

Similar to disruptive data poisoning, which reduces the salient features in a dataset, the associative poisoning attack is not reliant on a few well-crafted poisoned samples, but rather on the entire dataset distribution. As detection methods often focus on monitoring suspicious behaviour by singular samples, we believe the associative poisoning would be more challenging to detect, particularly since an adversary could elect to target arbitrary features, not just existing labels. We also demonstrate that, even with limited access to the training data, our attack can impact model behaviour, which could be leveraged by adversaries to introduce slight bias in datasets.

\subsection{Associative poisoning beyond generative models}
While not explored in this thesis, associative poisoning should not be limited to impacting unsupervised generative machine learning. The idea itself deals with the distribution of datasets and could likely be expanded to impact other models relying on deriving information from a dataset. Such models could include supervised and reinforcement learning models, unsupervised inference models such as clustering and autoencoders, as well as other statistical models, like regression.

For unsupervised generative models, the adversary's goal is to change the distribution of the generated data. The goal, and hence the selection of features, would align with the targeted model's goal, i.e., the bias introduced by the associative poisoning would need to align with the target model to actually matter. For example, in targeting classification, the adversary would need to identify and select features that are important to the classification task in order to introduce bias into the classification, e.g., associating a specific medical diagnosis with a specific gender.

\subsection{Possible Defences Against the Attack}
Because associative poisoning only tweaks the co‐occurrence of selected features while preserving each feature’s marginal distribution, defences examining the training set in isolation will likely fail. Effective mitigation must therefore draw on information beyond the corrupted dataset or impose strong assumptions about legitimate feature relationships.

We first examine other poisoning attacks and how these approaches align with the adversarial method, as well as how they would likely perform against associative poisoning. 

Some methods to detect backdoor attacks focus on observing anomalous behaviour in the model. Static model inspections search for anomalous connectivity or weight magnitudes, while dynamic methods monitor neuron activations during inference to identify abnormal responses~\cite{rawat2022devil}. Spectral-signature approaches cluster deep-layer activations to isolate suspicious sample subsets~\cite{tran2018spectral}, and activation‐based detectors flag inputs that trigger rare neuron patterns \cite{chen2018detecting, wang2020practical}. Adversarial neural pruning removes neurons associated with malicious correlations \cite{chou2023backdoor}. While effective against classic trigger-based attacks, these methods are ill-suited to associative poisoning, as the injected bias is diffuse, i.e., it is distributed across many samples, and would likely only produce activations similar to the original associations, making it challenging to differentiate between benign and adversarial associations without outside assumptions about the data.

Other methods rely on outlier detection, such as clustering-based filters and robust estimators, which assume that poisoned samples exhibit individually detectable anomalies~\cite{diakonikolas2019sever, paudice2018detection}. Associative poisoning, on the other hand, modifies only a single feature per sample, resulting in minimal deviation per sample. As a consequence, poisoned instances remain statistically consistent with the surrounding data, making them difficult to isolate using conventional sample-level outlier criteria. While some feature modifications could result in outlier artefacts, this would not be a flaw in the attack itself, but rather in the specific implementation of the modification.

To address the limitations of existing defences, we propose a two-stage mitigation strategy. First, pairwise association metrics (e.g., mutual information) are computed for all labelled feature pairs. Pairs whose scores lie in the extreme tails of the resulting distribution are flagged as suspicious. Second, independence is enforced for each flagged pair by modifying the training data through conditional resampling, targeted relabelling, or reverse associative poisoning. This process could be further strengthened by incorporating a small, trusted reference dataset or conducting a brief expert audit of the top-ranked anomalous pairs, thereby anchoring the detection to external ground truth prior to model training. There are, however, limitations to implementing such a scheme. First, it requires some kind of labels in order to calculate the distribution and reveal suspicious pairs, which for many datasets are challenging or impractical to produce. Second, is that enforcing independence on multiple pairs of features could heavily reduce the dataset when removing potentially poisoned samples, resulting in a less usable dataset. Third, there is the challenge of differentiating between benign and adversarial associations, as most datasets would contain a varying degree of benign associations, some of which could be stronger than an adversarial association.

\subsection{Limitations of the results}
The proposed associative poisoning attack provably alters the statistical relationship between chosen binary features in the training data, as well as for continuous features, with experiments confirming this effect, even when the adversary has limited control over the data. We also demonstrate that the poisoning causes marginal features to be retained. 

However, certain empirical observations warrant further discussion.

\paragraph{Reduced association shifts}
First, we observe that the magnitude of association shifts in the generated samples is often reduced compared to the manipulated training set, which we attribute to two factors.

First, distribution matching by generative models is inherently approximate. Architectures such as GANs and diffusion models aim to replicate the input distribution, but their unsupervised training objectives prioritise perceptual quality over the preservation of fine-grained statistical relationships, especially when certain feature categories are rare or under-represented. Moreover, these models are typically trained and evaluated on natural-image benchmarks. When applied to specialised domains, such as medical imaging or scientific data, performance often degrades, making it difficult to assess training quality. Evaluation metrics like FID rely on feature extractors that are pretrained on ImageNet, which may fail to capture domain-specific performance.

Second, to quantify changes in feature associations, we rely on automatic classification to label the features in generated samples. However, classification accuracy varies, and misclassifications introduce noise into our measures of MI and MCC. These measurement errors do not diminish the real-world impact of the attack, as they affect only the evaluation process. However, they can obscure the true magnitude of the induced association shifts. Using more accurate, domain-specific classifiers, along with careful calibration, would reduce this variance and yield more reliable measurements.

\paragraph{Counter-intuitive MI}
In Figures \ref{fig:mi_mcc_recipe1m}, we observe that, while the correlation increases from the positive attack, MI decreases, running counter to the hypothetical increase predicted by theory. We believe that the primary reason for this is that the features were originally negatively associated. As described in the theory in Section \ref{sec:theory_binary} and visualised in Figure \ref{fig:recipe1m_MI_MCC_vs_X}, to overcome this original association, the features would need to first reach independence. We can observe the same phenomenon in all negative attacks on continuous features in Figure \ref{fig:mi_pcc_celeba_dgan_continious}, since all average colours are strongly positively associated. For all these results, the corresponding correlation measure (which increases/decreases linearly instead) always aligns with the theoretical results.

\paragraph{Stealthiness and significant results}
We consider the stealthiness of the attack to be related to how ``detectable'' the modification introduced by the attack is. While some results indicate that MP, mean, and FID values differ between the clean and attacked models, in a realistic setting, this would not be enough for the attack to be detected. In a real scenario, the target would likely not have access to clean models as a comparison, and as such, the difference introduced by the adversary would need to be strong enough to arouse suspicion.

\subsection{More complex AP modification}
The implementation of the modification strategy is straightforward, involving simply searching for replacement pairs and identifying desired replacements from the hold-out set. 

For one, using the hold-out set places a limitation on the number of samples that can realistically be replaced (up to 20\% of the total dataset). Additionally, since the hold-out set is sampled from the same distribution as the training set, the real number of samples that could be replaced is further reduced, since the number of applicable samples would gradually decrease (in binary, the desired group would eventually empty, and for continuous the distance to the desired target would be too large), assuming that duplicates are disallowed.

More complex methods for modifying features could be implemented to overcome these limitations, e.g., drawing inspiration from DeepFake models to retain certain features while replacing others \cite{westerlund2019emergence}. Such approaches would, of course, come at their own cost, possibly in the form of detectable artefacts from the modification procedure, or high resource cost.

\section{Conclusion and future work}
\label{sec:conclusion_future}

We have demonstrated that generative machine learning is susceptible to a novel associative poisoning attack, which introduces malicious associations or removes existing associations between features in a dataset without affecting the rates at which the features appear. We provide a formal proof that the attack satisfies both fidelity and stealth requirements, and we empirically show that state-of-the-art generators will replicate these malicious associations while only minimally affecting feature prevalence and overall sample quality. We have further examined the limitations of existing defences and find that none are well-suited to detect or mitigate this type of attack. By exposing this vulnerability, we aim to highlight a broader class of threats that exploit subtle shifts in feature dependencies. These threats may compromise not only generative models but also any downstream system that relies on distributional assumptions for learning or inference.

Several important directions for future research remain. As our attack is agnostic to model architecture and data domain, it is essential to evaluate its effectiveness across a broader range of tasks, including text generation, audio synthesis, and non-image modalities such as tabular or graph data. The development of targeted defences against associative poisoning remains an open challenge. 

Finally, it would be valuable to explore how associative poisoning might be combined with other attack vectors, such as membership inference or model extraction, to construct more powerful, multi-faceted adversarial strategies.

\section*{Declaration of generative AI and AI-assisted technologies in the manuscript preparation process.}
During the preparation of this work, the authors used ChatGPT5 for reformulations to ensure clarity and suitable language. After using this tool/service, the authors reviewed and edited the content as needed and take full responsibility for the content of the published article.

\bibliographystyle{plainnat}
\bibliography{bibliography}

\begin{thebibliography}{52}
\providecommand{\natexlab}[1]{#1}
\providecommand{\url}[1]{\texttt{#1}}
\expandafter\ifx\csname urlstyle\endcsname\relax
  \providecommand{\doi}[1]{doi: #1}\else
  \providecommand{\doi}{doi: \begingroup \urlstyle{rm}\Url}\fi

\bibitem[Abay et~al.(2019)Abay, Zhou, Kantarcioglu, Thuraisingham, and Sweeney]{abay2019privacy}
Nazmiye~Ceren Abay, Yan Zhou, Murat Kantarcioglu, Bhavani Thuraisingham, and Latanya Sweeney.
\newblock Privacy preserving synthetic data release using deep learning.
\newblock In \emph{Machine Learning and Knowledge Discovery in Databases: European Conference, ECML PKDD 2018, Dublin, Ireland, September 10--14, 2018, Proceedings, Part I 18}, pages 510--526. Springer, 2019.

\bibitem[Barreno et~al.(2010)Barreno, Nelson, Joseph, and Tygar]{barreno2010security}
Marco Barreno, Blaine Nelson, Anthony~D Joseph, and J~Doug Tygar.
\newblock The security of machine learning.
\newblock \emph{Machine Learning}, 81:\penalty0 121--148, 2010.

\bibitem[Biggio et~al.(2011)Biggio, Nelson, and Laskov]{biggio2011support}
Battista Biggio, Blaine Nelson, and Pavel Laskov.
\newblock Support vector machines under adversarial label noise.
\newblock In \emph{Asian conference on machine learning}, pages 97--112. PMLR, 2011.

\bibitem[Biggio et~al.(2013)Biggio, Pillai, Rota~Bul{\`o}, Ariu, Pelillo, and Roli]{13}
Battista Biggio, Ignazio Pillai, Samuel Rota~Bul{\`o}, Davide Ariu, Marcello Pelillo, and Fabio Roli.
\newblock Is data clustering in adversarial settings secure?
\newblock In \emph{Proceedings of the 2013 ACM workshop on Artificial intelligence and security}, pages 87--98, 2013.

\bibitem[Borji(2022)]{borji2022generated}
Ali Borji.
\newblock Generated faces in the wild: Quantitative comparison of stable diffusion, midjourney and dall-e 2.
\newblock \emph{arXiv preprint arXiv:2210.00586}, 2022.

\bibitem[Caruana(1997)]{caruana1997multitask}
Rich Caruana.
\newblock Multitask learning.
\newblock \emph{Machine learning}, 28\penalty0 (1):\penalty0 41--75, 1997.

\bibitem[Chaudhury et~al.(2021)Chaudhury, Roy, Mishra, and Yamasaki]{R2_14}
Subhajit Chaudhury, Hiya Roy, Sourav Mishra, and T.~Yamasaki.
\newblock Adversarial training time attack against discriminative and generative convolutional models.
\newblock \emph{IEEE Access}, 9:\penalty0 109241--109259, 2021.
\newblock URL \url{https://api.semanticscholar.org/CorpusID:236980747}.

\bibitem[Chen et~al.(2018)Chen, Carvalho, Baracaldo, Ludwig, Edwards, Lee, Molloy, and Srivastava]{chen2018detecting}
Bryant Chen, Wilka Carvalho, Nathalie Baracaldo, Heiko Ludwig, Benjamin Edwards, Taesung Lee, Ian Molloy, and Biplav Srivastava.
\newblock Detecting backdoor attacks on deep neural networks by activation clustering.
\newblock \emph{arXiv preprint arXiv:1811.03728}, 2018.

\bibitem[Chou et~al.(2023)Chou, Chen, and Ho]{chou2023backdoor}
Sheng-Yen Chou, Pin-Yu Chen, and Tsung-Yi Ho.
\newblock How to backdoor diffusion models?
\newblock In \emph{Proceedings of the IEEE/CVF Conference on Computer Vision and Pattern Recognition}, pages 4015--4024, 2023.

\bibitem[Cin{\`a} et~al.(2023)Cin{\`a}, Grosse, Demontis, Vascon, Zellinger, Moser, Oprea, Biggio, Pelillo, and Roli]{cina2023wild}
Antonio~Emanuele Cin{\`a}, Kathrin Grosse, Ambra Demontis, Sebastiano Vascon, Werner Zellinger, Bernhard~A Moser, Alina Oprea, Battista Biggio, Marcello Pelillo, and Fabio Roli.
\newblock Wild patterns reloaded: A survey of machine learning security against training data poisoning.
\newblock \emph{ACM Computing Surveys}, 55\penalty0 (13s):\penalty0 1--39, 2023.

\bibitem[Creswell et~al.(2017)Creswell, Bharath, and Sengupta]{creswell2017latentpoison}
Antonia Creswell, Anil~A Bharath, and Biswa Sengupta.
\newblock Latentpoison-adversarial attacks on the latent space.
\newblock \emph{arXiv preprint arXiv:1711.02879}, 2017.

\bibitem[Diakonikolas et~al.(2019)Diakonikolas, Kamath, Kane, Li, Steinhardt, and Stewart]{diakonikolas2019sever}
Ilias Diakonikolas, Gautam Kamath, Daniel Kane, Jerry Li, Jacob Steinhardt, and Alistair Stewart.
\newblock Sever: A robust meta-algorithm for stochastic optimization.
\newblock In \emph{International Conference on Machine Learning}, pages 1596--1606. PMLR, 2019.

\bibitem[Ding et~al.(2019{\natexlab{a}})Ding, Tian, Xu, Li, and Zhong]{FW1_11}
Shaohua Ding, Yulong Tian, Fengyuan Xu, Qun Li, and Sheng Zhong.
\newblock Trojan attack on deep generative models in autonomous driving.
\newblock In \emph{International Conference on Security and Privacy in Communication Systems}, pages 299--318. Springer, 2019{\natexlab{a}}.

\bibitem[Ding et~al.(2019{\natexlab{b}})Ding, Tian, Xu, Li, and Zhong]{ding2019trojan}
Shaohua Ding, Yulong Tian, Fengyuan Xu, Qun Li, and Sheng Zhong.
\newblock Trojan attack on deep generative models in autonomous driving.
\newblock In \emph{Security and Privacy in Communication Networks: 15th EAI International Conference, SecureComm 2019, Orlando, FL, USA, October 23-25, 2019, Proceedings, Part I 15}, pages 299--318. Springer, 2019{\natexlab{b}}.

\bibitem[Duncan(1970)]{duncan1970calculation}
Tyrone~E Duncan.
\newblock On the calculation of mutual information.
\newblock \emph{SIAM Journal on Applied Mathematics}, 19\penalty0 (1):\penalty0 215--220, 1970.

\bibitem[Erickson et~al.(2020)Erickson, Mueller, Shirkov, Zhang, Larroy, Li, and Smola]{erickson2020autogluon}
Nick Erickson, Jonas Mueller, Alexander Shirkov, Hang Zhang, Pedro Larroy, Mu~Li, and Alexander Smola.
\newblock Autogluon-tabular: Robust and accurate automl for structured data.
\newblock \emph{arXiv preprint arXiv:2003.06505}, 2020.

\bibitem[Gretton et~al.(2007)Gretton, Fukumizu, Teo, Song, Sch{\"o}lkopf, and Smola]{gretton2007kernel}
Arthur Gretton, Kenji Fukumizu, Choon Teo, Le~Song, Bernhard Sch{\"o}lkopf, and Alex Smola.
\newblock A kernel statistical test of independence.
\newblock \emph{Advances in neural information processing systems}, 20, 2007.

\bibitem[Hendrycks et~al.(2018)Hendrycks, Mazeika, Wilson, and Gimpel]{hendrycks2018using}
Dan Hendrycks, Mantas Mazeika, Duncan Wilson, and Kevin Gimpel.
\newblock Using trusted data to train deep networks on labels corrupted by severe noise.
\newblock \emph{Advances in neural information processing systems}, 31, 2018.

\bibitem[Heusel et~al.(2017)Heusel, Ramsauer, Unterthiner, Nessler, and Hochreiter]{heusel2017gans}
Martin Heusel, Hubert Ramsauer, Thomas Unterthiner, Bernhard Nessler, and Sepp Hochreiter.
\newblock Gans trained by a two time-scale update rule converge to a local nash equilibrium.
\newblock \emph{Advances in neural information processing systems}, 30, 2017.

\bibitem[Hollander et~al.(2013)Hollander, Wolfe, and Chicken]{HollanderWolfeChicken2013}
Myles Hollander, Douglas~A. Wolfe, and Eric Chicken.
\newblock \emph{Nonparametric Statistical Methods}.
\newblock Wiley, 3 edition, 2013.

\bibitem[Kraskov et~al.(2004)Kraskov, St{\"o}gbauer, and Grassberger]{kraskov2004estimating}
Alexander Kraskov, Harald St{\"o}gbauer, and Peter Grassberger.
\newblock Estimating mutual information.
\newblock \emph{Physical Review E—Statistical, Nonlinear, and Soft Matter Physics}, 69\penalty0 (6):\penalty0 066138, 2004.

\bibitem[Lawrence and Lin(1989)]{lawrence1989concordance}
I~Lawrence and Kuei Lin.
\newblock A concordance correlation coefficient to evaluate reproducibility.
\newblock \emph{Biometrics}, pages 255--268, 1989.

\bibitem[Li et~al.(2017)Li, Yang, Song, Cao, Luo, and Li]{li2017learning}
Yuncheng Li, Jianchao Yang, Yale Song, Liangliang Cao, Jiebo Luo, and Li-Jia Li.
\newblock Learning from noisy labels with distillation.
\newblock In \emph{Proceedings of the IEEE international conference on computer vision}, pages 1910--1918, 2017.

\bibitem[Liu et~al.(2015)Liu, Luo, Wang, and Tang]{liu2015faceattributes}
Ziwei Liu, Ping Luo, Xiaogang Wang, and Xiaoou Tang.
\newblock Deep learning face attributes in the wild.
\newblock In \emph{Proceedings of International Conference on Computer Vision (ICCV)}, December 2015.

\bibitem[Mann and Whitney(1947)]{mann1947test}
Henry~B Mann and Donald~R Whitney.
\newblock On a test of whether one of two random variables is stochastically larger than the other.
\newblock \emph{The annals of mathematical statistics}, pages 50--60, 1947.

\bibitem[Matthews(1975)]{matthews1975comparison}
Brian~W Matthews.
\newblock Comparison of the predicted and observed secondary structure of t4 phage lysozyme.
\newblock \emph{Biochimica et Biophysica Acta (BBA)-Protein Structure}, 405\penalty0 (2):\penalty0 442--451, 1975.

\bibitem[McGill(1954)]{mcgill1954multivariate}
William McGill.
\newblock Multivariate information transmission.
\newblock \emph{Transactions of the IRE Professional Group on Information Theory}, 4\penalty0 (4):\penalty0 93--111, 1954.

\bibitem[Mei and Zhu(2015)]{mei2015using}
Shike Mei and Xiaojin Zhu.
\newblock Using machine teaching to identify optimal training-set attacks on machine learners.
\newblock In \emph{Proceedings of the {AAAI} conference on artificial intelligence}, 2015.

\bibitem[Ning et~al.(2023)Ning, Sangineto, Porrello, Calderara, and Cucchiara]{ning2023input}
Mang Ning, Enver Sangineto, Angelo Porrello, Simone Calderara, and Rita Cucchiara.
\newblock Input perturbation reduces exposure bias in diffusion models.
\newblock \emph{arXiv preprint arXiv:2301.11706}, 2023.

\bibitem[Nkashama et~al.(2022)Nkashama, Soltani, Verdier, Frappier, Tardif, and Kabanza]{nkashama2022robustness}
D~Nkashama, Arian Soltani, Jean-Charles Verdier, Marc Frappier, Pierre-Marting Tardif, and Froduald Kabanza.
\newblock Robustness evaluation of deep unsupervised learning algorithms for intrusion detection systems.
\newblock \emph{arXiv preprint arXiv:2207.03576}, 2022.

\bibitem[Paudice et~al.(2018)Paudice, Mu{\~n}oz-Gonz{\'a}lez, Gyorgy, and Lupu]{paudice2018detection}
Andrea Paudice, Luis Mu{\~n}oz-Gonz{\'a}lez, Andras Gyorgy, and Emil~C Lupu.
\newblock Detection of adversarial training examples in poisoning attacks through anomaly detection.
\newblock \emph{arXiv preprint arXiv:1802.03041}, 2018.

\bibitem[Pearson(1895)]{pearson1895vii}
Karl Pearson.
\newblock Vii. note on regression and inheritance in the case of two parents.
\newblock \emph{proceedings of the royal society of London}, 58\penalty0 (347-352):\penalty0 240--242, 1895.

\bibitem[Peri et~al.(2020)Peri, Gupta, Huang, Fowl, Zhu, Feizi, Goldstein, and Dickerson]{peri2020deep}
Neehar Peri, Neal Gupta, W~Ronny Huang, Liam Fowl, Chen Zhu, Soheil Feizi, Tom Goldstein, and John~P Dickerson.
\newblock Deep k-nn defense against clean-label data poisoning attacks.
\newblock In \emph{Computer Vision--ECCV 2020 Workshops: Glasgow, UK, August 23--28, 2020, Proceedings, Part I 16}, pages 55--70. Springer, 2020.

\bibitem[Rawat et~al.(2022)Rawat, Levacher, and Sinn]{rawat2022devil}
Ambrish Rawat, Killian Levacher, and Mathieu Sinn.
\newblock The devil is in the gan: backdoor attacks and defenses in deep generative models.
\newblock In \emph{European Symposium on Research in Computer Security}, pages 776--783. Springer, 2022.

\bibitem[Ross(2014)]{ross2014mutual}
Brian~C Ross.
\newblock Mutual information between discrete and continuous data sets.
\newblock \emph{PloS one}, 9\penalty0 (2):\penalty0 e87357, 2014.

\bibitem[Salem et~al.(2020)Salem, Sautter, Backes, Humbert, and Zhang]{salem2020baaan}
Ahmed Salem, Yannick Sautter, Michael Backes, Mathias Humbert, and Yang Zhang.
\newblock Baaan: Backdoor attacks against autoencoder and gan-based machine learning models.
\newblock \emph{arXiv preprint arXiv:2010.03007}, 2020.

\bibitem[Salvador et~al.(2017)Salvador, Hynes, Aytar, Marin, Ofli, Weber, and Torralba]{salvador2017learning}
Amaia Salvador, Nicholas Hynes, Yusuf Aytar, Javier Marin, Ferda Ofli, Ingmar Weber, and Antonio Torralba.
\newblock Learning cross-modal embeddings for cooking recipes and food images.
\newblock In \emph{Proceedings of the IEEE Conference on Computer Vision and Pattern Recognition}, 2017.

\bibitem[Shafahi et~al.(2018)Shafahi, Huang, Najibi, Suciu, Studer, Dumitras, and Goldstein]{shafahi2018poison}
Ali Shafahi, W~Ronny Huang, Mahyar Najibi, Octavian Suciu, Christoph Studer, Tudor Dumitras, and Tom Goldstein.
\newblock Poison frogs! targeted clean-label poisoning attacks on neural networks.
\newblock \emph{Advances in neural information processing systems}, 31, 2018.

\bibitem[Shannon(1948)]{shannon1948mathematical}
Claude~E Shannon.
\newblock A mathematical theory of communication.
\newblock \emph{The Bell system technical journal}, 27\penalty0 (3):\penalty0 379--423, 1948.

\bibitem[Singh et~al.(2023)Singh, Kumar, and Mehra]{singh2023chat}
Shashi~Kant Singh, Shubham Kumar, and Pawan~Singh Mehra.
\newblock Chat gpt \& google bard ai: A review.
\newblock In \emph{2023 International Conference on IoT, Communication and Automation Technology (ICICAT)}, pages 1--6. IEEE, 2023.

\bibitem[Souly et~al.(2025)Souly, Rando, Chapman, Davies, Hasircioglu, Shereen, Mougan, Mavroudis, Jones, Hicks, Carlini, Gal, and Kirk]{souly2025poisoningattacksllmsrequire}
Alexandra Souly, Javier Rando, Ed~Chapman, Xander Davies, Burak Hasircioglu, Ezzeldin Shereen, Carlos Mougan, Vasilios Mavroudis, Erik Jones, Chris Hicks, Nicholas Carlini, Yarin Gal, and Robert Kirk.
\newblock Poisoning attacks on llms require a near-constant number of poison samples, 2025.
\newblock URL \url{https://arxiv.org/abs/2510.07192}.

\bibitem[Sz{\'e}kely et~al.(2007)Sz{\'e}kely, Rizzo, and Bakirov]{szekely2007measuring}
G{\'a}bor~J Sz{\'e}kely, Maria~L Rizzo, and Nail~K Bakirov.
\newblock Measuring and testing dependence by correlation of distances.
\newblock 2007.

\bibitem[Tian et~al.(2022)Tian, Cui, Liang, and Yu]{tian2022comprehensive}
Zhiyi Tian, Lei Cui, Jie Liang, and Shui Yu.
\newblock A comprehensive survey on poisoning attacks and countermeasures in machine learning.
\newblock \emph{ACM Computing Surveys}, 55\penalty0 (8):\penalty0 1--35, 2022.

\bibitem[Tran et~al.(2018)Tran, Li, and Madry]{tran2018spectral}
Brandon Tran, Jerry Li, and Aleksander Madry.
\newblock Spectral signatures in backdoor attacks.
\newblock \emph{Advances in neural information processing systems}, 31, 2018.

\bibitem[Wang et~al.(2020)Wang, Zhang, Liu, Chen, Xiong, and Wang]{wang2020practical}
Ren Wang, Gaoyuan Zhang, Sijia Liu, Pin-Yu Chen, Jinjun Xiong, and Meng Wang.
\newblock Practical detection of trojan neural networks: Data-limited and data-free cases.
\newblock In \emph{Computer Vision--ECCV 2020: 16th European Conference, Glasgow, UK, August 23--28, 2020, Proceedings, Part XXIII 16}, pages 222--238. Springer, 2020.

\bibitem[Wang et~al.(2022{\natexlab{a}})Wang, Zheng, He, Chen, and Zhou]{wang2022diffusiongan}
Zhendong Wang, Huangjie Zheng, Pengcheng He, Weizhu Chen, and Mingyuan Zhou.
\newblock {Diffusion-GAN}: Training gans with diffusion.
\newblock \emph{arXiv preprint arXiv:2206.02262}, 2022{\natexlab{a}}.
\newblock URL \url{https://arxiv.org/abs/2206.02262}.

\bibitem[Wang et~al.(2022{\natexlab{b}})Wang, Ma, Wang, Hu, Qin, and Ren]{wang2022threats}
Zhibo Wang, Jingjing Ma, Xue Wang, Jiahui Hu, Zhan Qin, and Kui Ren.
\newblock Threats to training: A survey of poisoning attacks and defenses on machine learning systems.
\newblock \emph{ACM Computing Surveys}, 55\penalty0 (7):\penalty0 1--36, 2022{\natexlab{b}}.

\bibitem[Westerlund(2019)]{westerlund2019emergence}
Mika Westerlund.
\newblock The emergence of deepfake technology: A review.
\newblock \emph{Technology innovation management review}, 9\penalty0 (11), 2019.

\bibitem[Yule(1912)]{yule1912methods}
G~Udny Yule.
\newblock On the methods of measuring association between two attributes.
\newblock \emph{Journal of the Royal Statistical Society}, 75\penalty0 (6):\penalty0 579--652, 1912.

\bibitem[Zhang et~al.(2019)Zhang, Chen, Wu, Chen, and Yu]{24}
Jiale Zhang, Junjun Chen, Di~Wu, Bing Chen, and Shui Yu.
\newblock Poisoning attack in federated learning using generative adversarial nets.
\newblock In \emph{2019 18th IEEE International Conference On Trust, Security And Privacy In Computing And Communications/13th IEEE International Conference On Big Data Science And Engineering (TrustCom/BigDataSE)}, pages 374--380. IEEE, 2019.

\bibitem[Zhang and Zhu(2017)]{zhang2017game}
Rui Zhang and Quanyan Zhu.
\newblock A game-theoretic analysis of label flipping attacks on distributed support vector machines.
\newblock In \emph{2017 51st Annual Conference on Information Sciences and Systems (CISS)}, pages 1--6. IEEE, 2017.

\bibitem[Zhao et~al.(2017)Zhao, An, Gao, and Zhang]{zhao2017efficient}
Mengchen Zhao, Bo~An, Wei Gao, and Teng Zhang.
\newblock Efficient label contamination attacks against black-box learning models.
\newblock In \emph{IJCAI}, pages 3945--3951, 2017.

\end{thebibliography}

\end{document}